\documentclass{article}

\usepackage{microtype}
\usepackage{graphicx}
\usepackage{subfigure}
\usepackage{booktabs} 
\usepackage[table]{xcolor}
\usepackage{caption}
\usepackage{minitoc}

\definecolor{uclablue}{RGB}{39, 116, 174}
\definecolor{bigaired}{RGB}{156, 0, 0}

\usepackage[colorlinks,citecolor=uclablue, linkcolor=bigaired]{hyperref}

\usepackage[accepted]{icml2025}

\usepackage{amsmath}
\usepackage{amssymb}
\usepackage{mathtools}
\usepackage{amsthm}

\usepackage[capitalize,noabbrev]{cleveref}

\usepackage{amsmath,amsfonts,bm}

\DeclareMathOperator{\defeq}{\stackrel{\text{def}}{\;=\;}}

\newtheorem{theorem}{Theorem}
\newtheorem{lemma}[theorem]{Lemma}

\def\eqref#1{equation~\ref{#1}}

\def\1{\bm{1}}

\DeclareMathAlphabet{\mathsfit}{\encodingdefault}{\sfdefault}{m}{sl}
\SetMathAlphabet{\mathsfit}{bold}{\encodingdefault}{\sfdefault}{bx}{n}

\newcommand{\E}{\mathbb{E}}

\definecolor{lightblue}{RGB}{210, 220, 250}

\usepackage{multicol}
\usepackage{ulem} 
\usepackage{wrapfig}
\usepackage{algorithm}
\usepackage{algorithmic}
\usepackage{pifont}
\usepackage{enumitem}
\usepackage{amsmath}
\usepackage{amssymb}
\usepackage{listings}
\usepackage{multirow}
\usepackage{hyperref} 
\usepackage{textgreek}
\setlist[itemize]{noitemsep, topsep=0pt}

\definecolor{customyellow}{HTML}{FFFACD} 

\theoremstyle{plain}
\theoremstyle{definition}

\theoremstyle{remark}

\newcommand{\colorone}[1]{\cellcolor{blue!5}#1} 
\newcommand{\colortwo}[1]{\cellcolor{blue!10}#1} 
\newcommand{\colorthree}[1]{\cellcolor{blue!15}#1} 
\newcommand{\colorfour}[1]{\cellcolor{blue!20}#1} 
\newcommand{\colorfive}[1]{\cellcolor{blue!30}#1} 

\usepackage[textsize=tiny]{todonotes}

\icmltitlerunning{How to Synthesize Text Data without Model Collapse?}

\begin{document}


\twocolumn[
\icmltitle{How to Synthesize Text Data without Model Collapse?}



\icmlsetsymbol{equal}{*}

\begin{icmlauthorlist}
\icmlauthor{Xuekai Zhu}{sjtu,bigai}
\icmlauthor{Daixuan Cheng}{bigai}
\icmlauthor{Hengli Li}{bigai,pku}
\icmlauthor{Kaiyan Zhang}{thu}
\icmlauthor{Ermo Hua}{thu}
\icmlauthor{Xingtai Lv}{thu}
\icmlauthor{Ning Ding}{thu}
\icmlauthor{Zhouhan Lin$^{\dagger}$}{sjtu,ailab}
\icmlauthor{Zilong Zheng$^{\dagger}$}{bigai}
\icmlauthor{Bowen Zhou$^{\dagger}$}{thu,ailab}
\end{icmlauthorlist}

\icmlaffiliation{sjtu}{LUMIA Lab, Shanghai Jiao Tong University}
\icmlaffiliation{bigai}{State Key Laboratory of General Artificial Intelligence, BIGAI}
\icmlaffiliation{thu}{Department of Electronic Engineering, Tsinghua University}
\icmlaffiliation{pku}{Institute for Artificial Intelligence, Peking University}
\icmlaffiliation{ailab}{Shanghai Artificial Intelligence Laboratory}

\icmlcorrespondingauthor{Zhouhan Lin}{lin.zhouhan@gmail.com}
\icmlcorrespondingauthor{Zilong Zheng}{zlzheng@bigai.ai}
\icmlcorrespondingauthor{Bowen Zhou}{zhoubowen@tsinghua.edu.cn}

\icmlkeywords{Machine Learning, ICML}

\vskip 0.3in
]



\printAffiliationsAndNotice{}  

\begin{abstract}
Model collapse in synthetic data indicates that iterative training on self-generated data leads to a gradual decline in performance. With the proliferation of AI models, synthetic data will fundamentally reshape the web data ecosystem. Future GPT-$\{n\}$ models will inevitably be trained on a blend of synthetic and human-produced data. In this paper, we focus on two questions: what is the impact of synthetic data on language model training, and how to synthesize data without model collapse? We first pre-train language models across different proportions of synthetic data, revealing a negative correlation between the proportion of synthetic data and model performance. We further conduct statistical analysis on synthetic data to uncover distributional shift phenomenon and over-concentration of n-gram features. Inspired by the above findings, we propose token editing on human-produced data to obtain semi-synthetic data. As a proof of concept, we theoretically demonstrate that token-level editing can prevent model collapse, as the test error is constrained by a finite upper bound. We conduct extensive experiments on pre-training from scratch, continual pre-training, and supervised fine-tuning. The results validate our theoretical proof that token-level editing improves model performance.
\end{abstract}

\begin{figure*}[t]
\small
  \centering
  \includegraphics[width=0.95\linewidth]{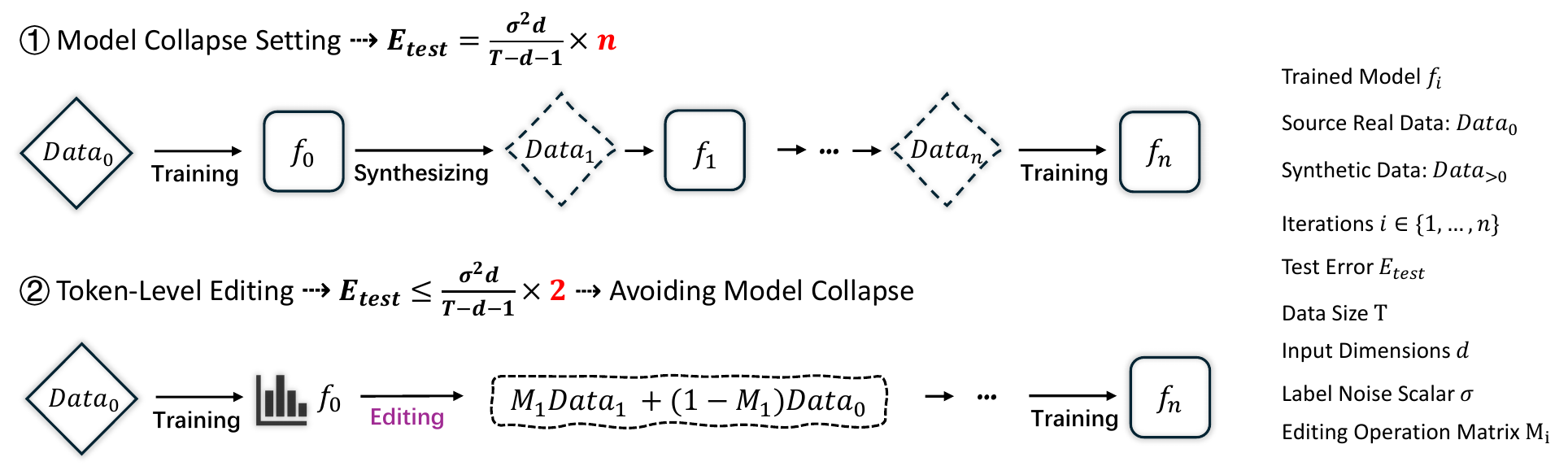}
  \vspace{-1em}
  \caption{Model collapse of synthetic data.~\textbf{\ding{172}} The model continuously trains on its previously generated data, leading to a gradual decline in model performance, i.e., model collapse. Starting from real data $Data_0$, the test error $E_{test}$ increases as $f_0$ undergoes iterative training on synthetic data $Data_{>0}$.
  \textbf{\ding{173}} ToEdit~(ours), we use a trained model for token-level editing rather than purely synthesizing data. 
  Leveraging $f_0$ and an operation matrix $M_i$ to edit the data, the test error is constrained within a fixed upper bound. Therefore, we can preserve the distribution coverage to avoid model collapse.  
  }
  \label{fig:main_figure}
  \vspace{-1em}
\end{figure*}

\section{Introduction}
As generative artificial intelligence (AI)~\citep{rombach2021highresolution,achiam2023gpt} becomes increasingly prevalent in research and industry, synthetic data will proliferate throughout the web data ecosystem. Consequently, future training of GPT-$\{n\}$ on a mixture of synthetic and human-produced data will be inevitable. Thus, model collapse is a critical concern that must be considered when training models on synthetic data. 

Model collapse refers to a degenerative process in which the output data of learned generative models contaminates the training sets of subsequent generations. As shown in Figure~\ref{fig:main_figure},
iterative training coupled with data synthesis induces a progressive accumulation of test errors~\citep{shumailov2024ai,dohmatob2024model}.
Consequently, generative models increasingly overfit to synthetic data distributions, failing to capture the complexity in human-produced data. Through successive iterations in Figure~\ref{fig:main_figure}, these distortions accumulate, finally undermining the model’s capacity. 

Recent studies focus on two aspects. First, theoretical foundations of model collapse. 
\citet{shumailov2024ai} and \citet{dohmatob2024model} identify the model collapse phenomenon and formalize a theoretical framework based on linear regression. \citet{gerstgrasser2024model} demonstrate that if synthetic data is accumulated while retaining the initial real data, the test error will be bounded, thus breaking model collapse. \citet{dohmatob2024tale,dohmatob2024strong} indicate that missing long tails of synthetic data lead to scaling law cutoff.
Second, practical implementations on synthetic datasets by diverse prompting.
Synthetic datasets~\citep{AlphaGeometryTrinh2024,zhang2024ultramedical} have been proven to boost the capabilities of language models. \citet{Cheng2024InstructionPL,cheng2024adapting, Maini2024RephrasingTW} rephrase text into more formal styles, thereby improving the data quality. There are still two key questions that require further investigation: \textbf{(Q1)} What is the impact of synthetic data on language model training?
\textbf{(Q2)} How can data be synthesized without causing model collapse?

In this paper, we address the first question by training language models on varying mixtures of synthetic and human-produced data, demonstrating non-iterative model collapse. Unlike the original model collapse setting which iteratively trains on self-generated data, we directly mix synthetic and human-produced data to create training datasets with different mixing ratios. The results show a negative correlation between performance and the proportion of synthetic data. Subsequent statistical analysis on distributions and features indicates coverage narrowing---synthetic data covers only a small portion of the human-produced data distribution---and over-concentration of synthetic n-gram features.
Based on the above findings, we address the second question by proposing \textbf{to}ken-level \textbf{edit}ing~(ToEdit), resamples and replaces data points with relatively high model confidence. As illustrated in Figure~\ref{fig:main_figure}, ToEdit preserves distribution coverage and theoretically constrains test error within a fixed upper bound. Extensive experiments across pre-training from scratch, continual pre-training, and supervised fine-tuning confirm its positive impact on model performance.\footnote{Work done during internship at BIGAI.}

\textbf{Contributions.} We summarize the key contributions of this work as follows\footnote{Code repository available at \url{https://github.com/Xuekai-Zhu/toedit}.}:
\begin{itemize}[leftmargin=*, topsep=0pt, noitemsep]
    \item We demonstrate non-iterative model collapse by pre-training language models on a mixture of synthetic and human-produced data~(\S~\ref{sec:non_iterative_model_collapse}): directly mixing pure synthetic data,
    without iterative training, leads to performance degradation. Furthermore, we perform a distributional statistical analysis, revealing that synthetic data leads to coverage narrowing and over-concentration of n-gram features. Even subsequent data selection struggles to correct the distribution (\S~\ref{sec:fail_in_language_model_pre-training}).
    \item We propose token-level editing with a theoretical proof to prevent model collapse~(\S~\ref{sec:token_level_data_editing}) and validate its effectiveness through experiments spanning pre-training from scratch, continual pre-training, and supervised fine-tuning of language models~(\S~\ref{sec:experiments}).
\end{itemize}

\section{Background}
\citet{shumailov2024ai,dohmatob2024model,dohmatob2024tale} 
demonstrate AI models trained recursively on data generated by earlier versions of themselves can result in performance degradation, ultimately rendering the AI model completely useless. This process can be formulated as follows:
\begin{small}
\begin{align*}
E_{test}(\hat{w}_{n+1}) = \frac{\sigma^2d}{T - d - 1}\times n.
\end{align*}
\end{small}
This indicates that the error will continuously increase with the number of iterations $n$. The detailed theoretical notation is provided in \S~\ref{sec:Theoretical_Analysis}.
\citet{dohmatob2024tale} further point out that synthetic data also contribute to a truncation of the scaling law. 
\citet{gerstgrasser2024model,seddik2024bad} further adjust the data iteration setting to data accumulation or real data mixing. They demonstrate that data accumulation can prevent model collapse. Inspired by the above work, we further explore the impact of synthetic data in pre-training and analyze its differences from real data. Building on our findings, we propose token editing as a method to prevent model collapse during data synthesis. Further comparisons are in Appendix~\ref{sec:more_related_work} and~\ref{sec:Explanation_of_Synthetic_Data}.

\begin{table*}[t]
\centering
\caption{Subdomain PPL evaluation results for GPT-2 Small (124M) pre-trained on data mixture. The PPL increases as the proportion of synthetic data grows, providing further confirmation of Figure~\ref{fig:gpt2-pretraining-and-eval}. 
}\label{tab:ppl_results_of_pile}
\resizebox{\textwidth}{!}{
\begin{tabular}{l|ccccccccccc|c}
        \toprule
        \textbf{} & \textbf{ArXiv} & \textbf{Books2} & \textbf{Books3} & \textbf{Math} & \textbf{Enron} & \textbf{EuroParl} & \textbf{FreeLaw} & \textbf{GitHub} & \textbf{PG-19} & \textbf{HackerNews} & \textbf{NIH} & \textbf{Avg} \\ 
        \midrule
        Human data         & 22.26 & 25.39 & 22.87 & 10.84 & 23.50 & 30.73 & 12.04 & 4.15 & 16.88 & 32.54 & 23.53 & \colorone{20.99} \\ 
        25\% Synthetic Data & 21.86 & 26.32 & 23.87 & 11.05 & 24.85 & 35.02 & 12.84 & 4.35 & 17.99 & 33.80 & 23.76 & \colortwo{22.06} \\ 
        50\% Synthetic Data & 22.50 & 28.01 & 25.75 & 10.84 & 26.56 & 41.99 & 14.02 & 4.67 & 19.70 & 36.12 & 24.61 & \colorthree{23.48} \\ 
        75\% Synthetic Data & 24.35 & 31.19 & 28.98 & 11.81 & 30.30 & 56.32 & 16.03 & 5.30 & 22.75 & 40.44 & 26.19 & \colorfour{27.60} \\ 
        Synthetic Data     & 35.60 & 43.72 & 47.72 & 17.25 & 66.97 & 129.75 & 29.62 & 12.00 & 50.14 & 87.95 & 39.48 & \colorfive{51.93} \\ 
        \midrule
        \textbf{} & \textbf{OpenSubts} & \textbf{OWT2} & \textbf{Phil} & \textbf{Pile-CC} & \textbf{PubMed-A} & \textbf{PubMed-C} & \textbf{StackEx} & \textbf{Ubuntu} & \textbf{USPTO} & \textbf{Wikipedia} & \textbf{Youtube} & \textbf{Avg} \\ 
        \midrule
        Human data         & 28.08 & 25.77 & 33.56 & 26.78 & 18.97 & 15.49 & 10.81 & 20.86 & 19.32 & 24.31 & 21.54 & \colorone{22.59} \\ 
        25\% Synthetic Data & 29.25 & 26.94 & 34.63 & 27.83 & 19.55 & 15.38 & 11.03 & 22.32 & 19.58 & 25.88 & 22.63 & \colortwo{23.91} \\ 
        50\% Synthetic Data & 31.00 & 28.76 & 37.48 & 29.36 & 20.51 & 15.89 & 11.54 & 23.53 & 20.51 & 27.57 & 24.91 & \colorthree{25.09} \\ 
        75\% Synthetic Data & 34.18 & 32.04 & 42.39 & 32.17 & 22.33 & 16.92 & 12.55 & 26.54 & 22.21 & 30.68 & 28.98 & \colorfour{28.64} \\ 
        Synthetic Data     & 57.83 & 53.94 & 78.18 & 54.69 & 34.82 & 23.87 & 20.47 & 51.78 & 37.24 & 46.12 & 65.49 & \colorfive{47.87} \\ 
        \bottomrule
    \end{tabular}}
\end{table*}

\begin{figure*}[t!]
\vspace{-1em}
\small
  \centering
  \includegraphics[width=\linewidth]{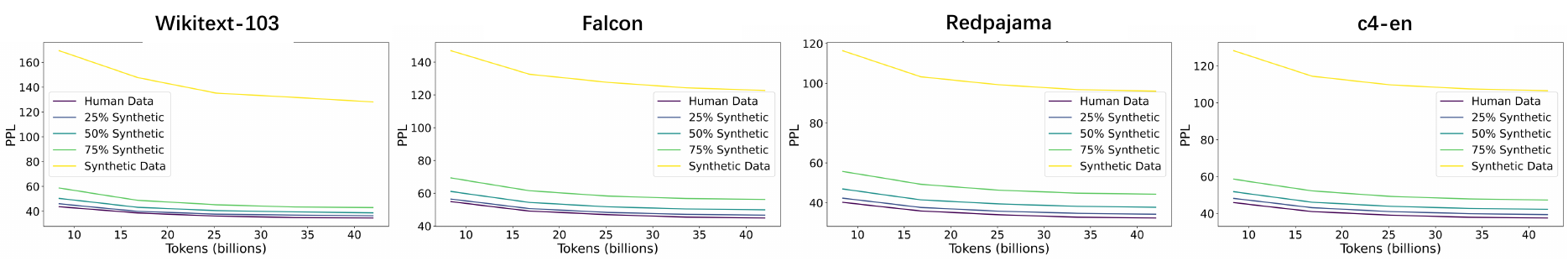}
  \vspace{-2em}
  \caption{Non-iterative model collapse. Training language models from scratch on AI-synthesized data or a mixture of human and synthetic data leads to performance degradation. This degradation is negatively correlated with the proportion of synthetic data used in training.
  \textit{Setting:} We pre-train GPT-2 Small (124M) on human data (Dolma~\citep{dolma}) and synthetic data (Cosmopedia~\citep{benallal2024cosmopedia}) and evaluate the PPL on the Paloma benchmark~\cite{Magnusson2023PalomaAB}. Training loss in Figure~\ref{fig:training_loss_synthetic_data}. Further validations on 22 subdomains and general downstream tasks are presented in Table~\ref{tab:ppl_results_of_pile} and Table~\ref{tab:human_vs_synthetic_downstream_tasks}, respectively.
  } 
  \label{fig:gpt2-pretraining-and-eval}
  \vspace{-1em}
\end{figure*}

\section{Non-Iterative Model Collapse}
Prior studies~\citep{shumailov2024ai,dohmatob2024model} investigate the curse of recursion, where iterative training on self-generated data leads to a degenerative process known as iterative model collapse. However, we often face direct data mixing of human-produced and synthetic data, and pre-training from scratch. We attempt to analyze model collapse in this more general scenario, called non-iterative model collapse. Specifically, we conduct pre-training on synthetic data mixtures and explore the reasons behind non-iterative model collapse through data distribution and characteristics. A more detailed comparison of iterative and non-iterative settings is in Appendix~\ref{sec:Non-Iterative-Model-Collapse}.

\subsection{Pre-training on Data Mixture}\label{sec:non_iterative_model_collapse}
In this section, we investigate the impact of synthetic data on pre-training. Compared with studies on SFT and RLHF, we examine synthetic data integration in a more fundamental stage of the language model. 

\underline{\textit{Setup}}\quad We pre-train GPT-2~\citep{radford2019language} and OLMo~\citep{OLMo} from scratch, using data mixtures containing 50B tokens each. We define the mixing ratio between human-produced and synthetic data as \(\alpha\), where \(0 \leq \alpha \leq 1\). The total amount of training data \(D_{\text{total}}\) is a combination of human-produced data \(D_{\text{human}}\) and synthetic data \(D_{\text{synthetic}}\), represented by the formula: 
$D_{\text{total}} = \alpha D_{\text{human}} + (1 - \alpha) D_{\text{synthetic}}$. We use Dolma~\citep{dolma} as source human-produced data. We use Cosmopedia~\citep{benallal2024cosmopedia} as the source synthetic data, which is distilled from \texttt{Mixtral-8x7B-Instruct-v0.1}~\citep{jiang2024mixtral}. 
To ensure rigorous validation and prevent data leakage, we construct a three-tier evaluation: (1) the Paloma benchmark~\citep{Magnusson2023PalomaAB}, including carefully decontaminated test sets for Dolma; (2) comprehensive PPL evaluation across 22 subdomains from the Pile~\citep{gao2020pile}; and (3) seven general downstream tasks as outlined in~\citep{Maini2024RephrasingTW}.


\textbf{Finding I: Incorporating synthetic data harms the language models pre-training.}\quad 
PPL results of Paloma benchmark and 22 subdomains are presented in Figure~\ref{fig:gpt2-pretraining-and-eval} and Table~\ref{tab:ppl_results_of_pile}, respectively. These results demonstrate that PPL on real-world validation sets increases as the proportion of synthetic data grows, indicating degraded model performance. When training from scratch, synthetic data does not benefit the model and may even hinder its learning process. However, incorporating human-produced data into the training mixture mitigates model collapse to some extent. Further results on general downstream tasks in Table~\ref{tab:human_vs_synthetic_downstream_tasks} and~\ref{tab:human_vs_synthetic_downstream_tasks_olmo} also corroborate the above findings. The overall trend shows a decline as the proportion of synthetic data increases, with models trained on purely synthetic data performing the worst. Compared to previous research on iterative model collapse~\citep{shumailov2024ai,dohmatob2024model,dohmatob2024tale}, the non-iterative damage caused by synthetic data is more concerning and directly relevant to the training of next-generation language models.
\begin{figure*}[t!]
\small
  \centering
  \includegraphics[width=\linewidth]{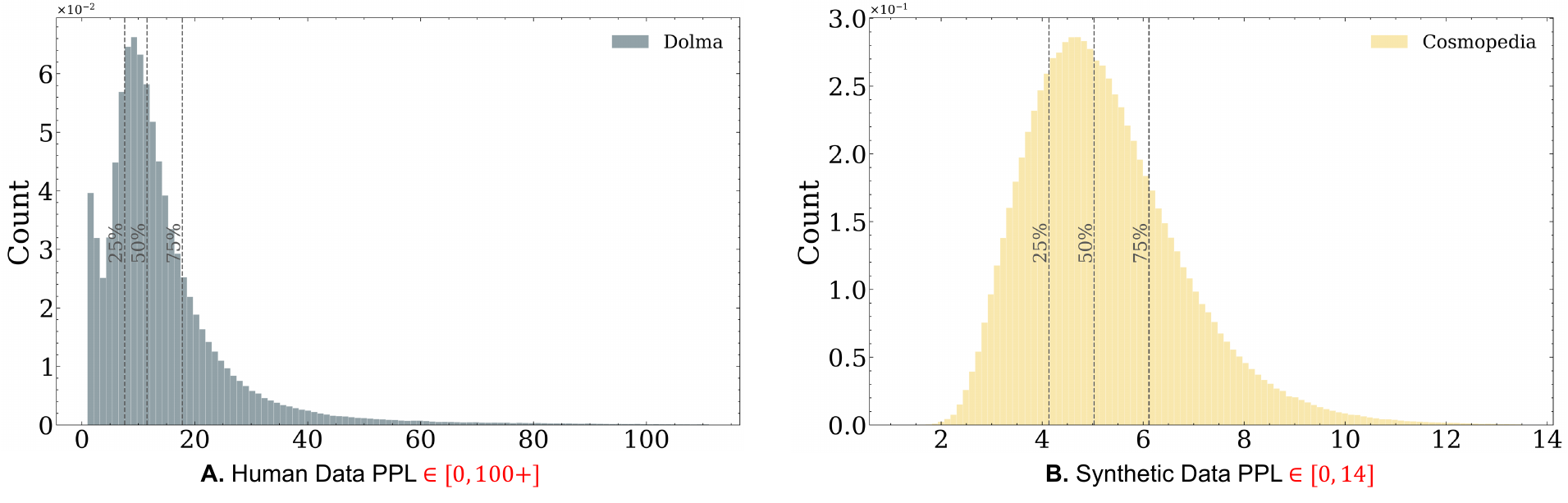}
  \vspace{-3em}
  \caption{PPL distribution of human and synthetic data estimated by Llama-3-8B. The synthetic data lacks the long tail of the human-produced data and is also concentrated within the first $25\%$ of the human-produced data distribution. \textbf{A.} Distribution of human-produced data is sharp with a long tail, spanning a wide range from 0 to over 100. \textbf{B.} The values are concentrated within a much narrower range, mostly between 0 and 14.
  The same trend estimated by StableLM-3B is demonstrated in Figure~\ref{fig:ppl_StabLM-Zephyr-3B}.
  } 
  \label{fig:ppl_distribution_llama_8b_lm}
\end{figure*}
\subsection{Why Does Synthetic Data Fail in Pre-training?}~\label{sec:fail_in_language_model_pre-training}
We conduct three statistical analyses: (1) sample-level distribution, (2) feature-based overlap, and (3) distribution-reference data selection. 
The experimental results reveal that, compared to human-produced data, synthetic data lacks the long-tail samples and suffers from coverage narrowing. 
The limited diversity and concentrated features in synthetic data make using human-produced data as a reference to select synthetic data particularly challenging.

\underline{\textit{Setup}}\quad We conduct statistical and feature-based analyses to explore why synthetic data fails in pre-training. (1) We leverage a prior distribution to estimate the human-produced and synthetic data. We use Llama-3-8B~\citep{llama3modelcard} and StableLM-Zephyr-3B~\citep{bellagente2024stable}. Different priors consistently yield the same results. (2) We analyze the n-gram features of human-produced and synthetic data from a feature-based perspective, such as n-gram response values. (3) Based on the features of human-produced data, we apply importance sampling~\cite{xie2023data} to filter synthetic data that closely aligns with human-produced features. More details of importance sampling are in \S~\ref{sec:dsir}. 

\textbf{Finding II: Synthetic data distribution not only lacks long tails but also exhibits significant coverage narrowing.}\quad Figure~\ref{fig:ppl_distribution_llama_8b_lm} illustrate that the PPL of synthetic data is confined to the lower $25\%$ of the human-produced data, failing to capture the full range and complexity of human-produced data distributions. Specifically, as illustrated in Figure~\ref{fig:ppl_distribution_llama_8b_lm}A, human-produced data exhibit a wide distribution in the range $[1, 100+]$, characterized by a sharp peak and a pronounced long tail. In contrast, as shown in Figure~\ref{fig:ppl_distribution_llama_8b_lm}B, the synthetic data is confined to a narrower range of $[0, 14]$, displaying a smoother distribution. Additional results of StabLM are shown in Figure~\ref{fig:ppl_StabLM-Zephyr-3B}. While the absolute PPL ranges estimated by different models may vary, the relative shapes and proportional ranges of these two distributions remain consistent. This phenomenon demonstrate that when scaling up to larger synthetic datasets, there will be a notable absence of the long tail, leading to severe coverage narrowing.
This limited coverage reduces the generalization ability and contribute to model collapse.



\begin{figure*}[t]
  \centering
  \includegraphics[width=\linewidth]{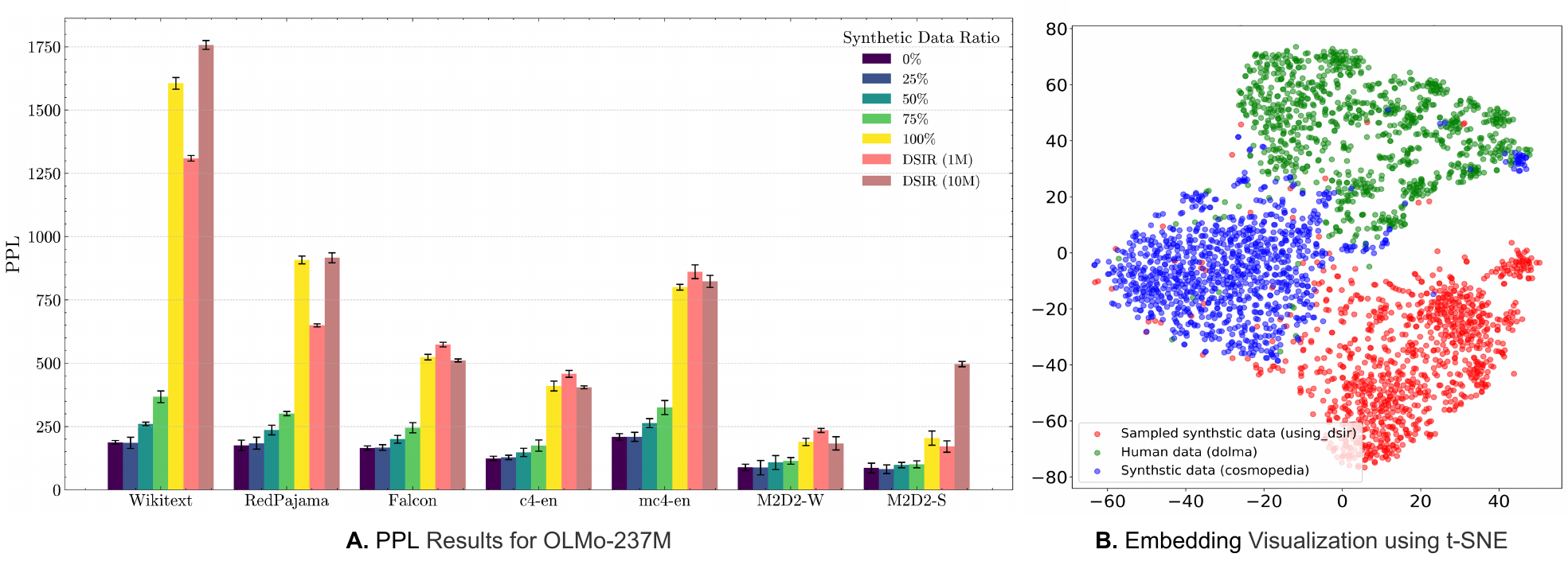}
  \vspace{-25pt}
  \caption{\textbf{A.} Pre-training results for selected synthetic data and other data mixtures on OLMo-237M. \textbf{B.} Embedding visualization  between human-produced, synthetic, and DSIR-selected data using sentence-transformer.} 
  \label{fig:tsne_and_dsir_ppl}
\end{figure*}

\textbf{Finding III: Synthetic data over-concentrates N-gram features.}\quad Based on the above distribution estimate, we further analyze why synthetic data fails at the feature level. 
Figure~\ref{fig:top_40_n_gram} and~\ref{fig:top_64_n_gram} demonstrate that synthetic data exhibits higher frequencies of certain bi-grams compared to human-produced data. To further examine feature-level differences, we hash uni-gram and bi-gram features into 10,000 hash buckets. As illustrated in Figure~\ref{fig:log_feature_diffence}, human-produced data displays a noticeably broader response range, while synthetic data features are concentrated in a few specific buckets. This indirectly supports our earlier observation of feature over-concentration. We then expanded the hash bucket range to 1,000 × 20,000 buckets and used a locality-sensitive hashing method to differentiate the features more precisely. The results remain consistent. As shown in Figure~\ref{fig:density_sampling}, most response values are near zero. Distinguishing features in synthetic data remains challenging.

\textbf{Finding IV: Distribution shifting cannot be mitigated through data selection.}\quad Inspired by recent data selection works~\citep{xie2023data,albalak2024survey}, we try to leverage the human-produced data features as a reference distribution for selecting synthetic samples. We apply importance sampling from DSIR~\citep{xie2023data} to filter synthetic data. As illustrated in Figure~\ref{fig:tsne_and_dsir_ppl}A, the training results of selected synthetic samples still fluctuates around the original performance of the synthetic data, indicating that even biased sampling cannot correct the distributional shift.
As shown in Figure~\ref{fig:tsne_and_dsir_ppl}B, the sampled data still fails to align with human-produced data in the embedding space, even at the boundary regions of the synthetic data. 

\subsection{Proposed Strategy}
Following these lessons so far, due to the coverage and feature over-concentration properties of synthetic data, the best approach is to rely entirely on human-produced data and avoid including synthetic data. However, we are still wondering how synthetic data can be used to enhance human-produced data. This leads to a general guideline for synthetic data: relying solely on synthetic data leads to model collapse, so preserving the primary human-produced data distribution is essential.
In that case, we propose token-level editing, which leverages a prior distribution to adjust the data. Our method can maintain the source distribution while improving the source data, called semi-synthetic data.
\begin{figure}
    \vspace{-1em}
    \centering
    \includegraphics[width=0.8\linewidth]{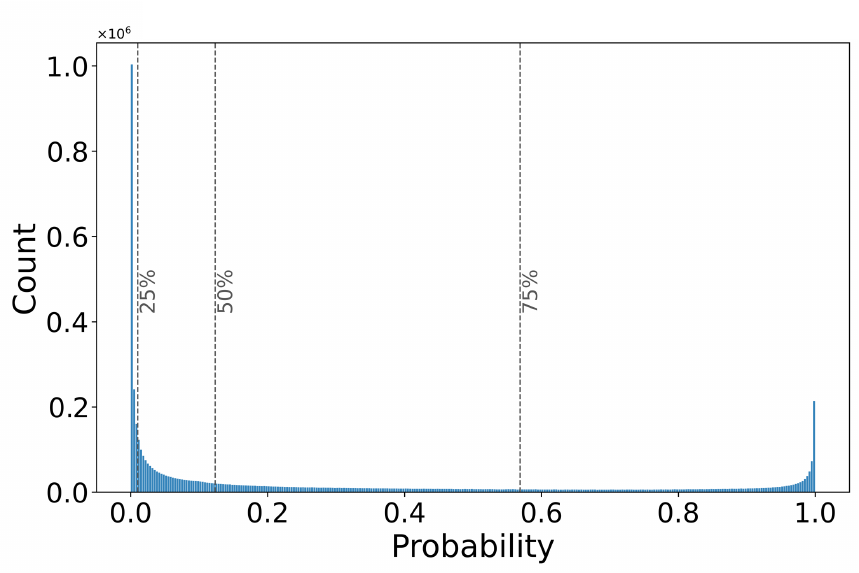}
    \vspace{-1em}
    \caption{U-shape token probability distribution of Dolma-sampled V6 estimated by Qwen-0.5B-Instruct~\citep{qwen2}.}
    \label{fig:qwen2_probs}
    \vspace{-1em}
\end{figure}

\section{Token-Level Editing}~\label{sec:token_level_data_editing}We introduce token-level editing as a method for generating semi-synthetic data. Furthermore, we present a theoretical analysis and proof demonstrating that the test squared error of our method has a finite upper bound, regardless of the number of iterations. This ensures prevention of model collapse while enhancing performance.
\subsection{Method}
We formulate data synthesis as follows: assuming $P$  is a prior distribution, given a sequence of tokens $\textbf{x} = (x_1, \dots, x_t)$, the full synthetic data is $\textbf{y} = (y_1, \dots, y_n)$. The synthesis process is derived as:
\begin{align}
P(y_1, \dots, y_n \mid \textbf{x}) = \prod_{i=1}^{n} P(y_i \mid \dots, y_{<i}, \textbf{x}).
\end{align}
This conditional probability formulation outlines the generation of synthetic data conditioned on the given token sequence. Then the synthetic data is used to train models. 

Inspired by previous studies of data selection~\citep{mindermann2022prioritized,ankner2024perplexed,lin2024rho}, prior distributions can serve as pointers indicating useless or learnable samples. In this case, we use a pre-trained language model to infer the pre-training corpus. As illustrated in Figure~\ref{fig:qwen2_probs}, even a model pre-trained on trillions of tokens can not fit the pre-training corpus perfectly. Specifically, $75\%$ is under 0.6 probability. The tokens with both high and low probabilities are the most concentrated, suggesting the potential for data filtering. We leverage this U-shape distribution as a pointer to resample tokens. Specifically, we use a language model as prior distribution to compute each token's conditional probability $P(\cdot|\textbf{x})$ if the probability exceeds a certain threshold \(P(\cdot|\textbf{x}) \geq p\), it indicates that the token is easy to learn, and we proceed with resampling at that point. The filtering potential of the U-shaped distribution is discussed in \S~\ref{sec:u_shape_distribution}.

Token-level Editing doesn't generate the entire sequence but leverages conditional probability $P(x_i \mid x_1, \dots, x_{i-1})$ to revise the input sequence. In this way, we can avoid using purely synthetic data while modifying the dataset to preserve long-tail features of human-produced data, aiming to obtain higher-quality semi-synthetic data. Token-level Editing can be formulated as follows:
\begin{align}
x_i' =
\begin{cases}
x_i, & \text{if } P(x_i \mid x_1, \dots, x_{i-1}) < p \\
\tilde{x}_i, & \text{if } P(x_i \mid x_1, \dots, x{i-1}) \geq p
\end{cases}
\end{align}

Where \(x_i'\) is the final token in the edited sequence. \(\tilde{x}_i\) is a token resampled from a prior distribution. We can adjust the threshold $p$ to balance retaining the structure of human-produced data while avoiding overfitting to synthetic data.

\subsection{Theoretical Analysis}\label{sec:Theoretical_Analysis}
To gain deeper mathematical insights, we utilize an analytical framework of the linear model and adopt notations in prior research~\citep{mobahi2020self,dohmatob2024model,gerstgrasser2024model}. This theoretical framework primarily considers a linear model that iteratively trains on its own generated data, similar to pipelines like self-play and self-distillation, but without complex constraints. The process involves training continuously on the data generated by the previous generation of the model. \cite{dohmatob2024model} point out that with iterative training, test errors accumulate progressively, eventually leading to model collapse. Building on this theoretical framework, we integrate our proposed token editing method and analyze whether our method can prevent model collapse.

\textbf{Notation and Preliminaries}
For a given distribution $P_{\Sigma,w,\sigma^2}$, the data $(x,y) \sim P_{\Sigma,w,\sigma^2}$ on $\mathbb{R}^d \times \mathbb{R}$, where $x$ is drawn from a multivariate normal distribution $x \sim \mathcal{N}(0, \Sigma)$, $\epsilon$ is an independent noise term sampled from $\mathcal{N}(0, \sigma^2)$, and the label $y$ is given by the linear model $y = x \cdot w^* + \epsilon$. 

\textbf{Iterative Data Editing Process}
We utilize the model obtained from the previous round of training to make a limited number of modifications. Specifically, we resample and replace data points with relatively high confidence. The editing operations are defined by the matrices $\{M_1, M_2, \dots, M_n\}$. The iterative data synthesis and model-fitting process is formalized as follows: 
\begin{align*}
P_{\Sigma,w^*,\sigma^2} \to P_{\Sigma,\hat{w}_1,\sigma^2} \to \ldots \to P_{\Sigma,\hat{w}_n,\sigma^2},
\end{align*}
where $n$ is the number of iterations. The detailed process of data editing and iterations is described as follows:

For \( n = 1 \), we begin by initializing the covariates/features as \( \tilde{X}_1 = X \). The target values are defined as \( \tilde{Y}_1 = \hat{Y}_1 = Xw^* + E_1 \), where \( E_1 \sim \mathcal{N}(0, \sigma^2 I_T) \). The linear model is then fitted by solving for \( \hat{w}_1 = \tilde{X}_1^{\dagger} \tilde{Y}_1 \). To proceed to the next iteration, we resample the data, obtaining \( \hat{Y}_2 = X \hat{w}_1 + E_2 \), with \( E_2 \sim \mathcal{N}(0, \sigma^2 I_T) \).

For \( n \geq 2 \), the input covariates/features remain as \( \tilde{X}_n^\top = X \), while the target values are updated using the edited targets, following the equation \( \tilde{Y}_n^\top = M_{n-1} \hat{Y}_n + (1 - M_{n-1}) \tilde{Y}_{n-1} \). The linear model is then fitted by computing \( \hat{w}_n = \tilde{X}_n^{\dagger} \tilde{Y}_n \). Finally, the data is resampled for the next iteration, yielding \( \hat{Y}_{n+1} = X \hat{w}_n + E_{n+1} \), where \( E_{n+1} \sim \mathcal{N}(0, \sigma^2 I_T) \).

The matrix $M_n$ is a diagonal matrix, where some diagonal elements are 1, while others are 0. The multiplication by M can be interpreted as an operation that selectively modifies certain data points (those corresponding to 1s) while retaining others (those corresponding to 0s). Then, the data editing process can be formulated as follows:
\begin{align}
    \tilde{Y}_n^\top = M_{n-1}\hat{Y}_{n} + (1-M_{n-1})\tilde{Y}_{n-1}
\end{align}
where \(\tilde{Y}_{n-1}\) is the data after editing in the $n-1$ generation, and \(\hat{Y}_{n}\) is the synthetic data from the \(n\)-th generation. This process can be described as: firstly, synthesizing labels for all inputs; secondly, the \(M\) matrix determining which data is edited and which is retained. For a matrix \( A \) with full column rank, its Moore-Penrose pseudo-inverse is \( A^+ = (A^\top A)^{-1} A^\top \). The noise terms \( E_1, E_2, \dots, E_n \) are independent of each other and the covariates/features. Since \( X \) has full column rank, \( \tilde{X}_n \) retains this property for all \( n \geq 1 \). 
\paragraph{Test Error}  Model collapse is ultimately reflected through test error. Following previous work, we adopt the standard test error definition as presented in~\citep{gerstgrasser2024model}. For any linear estimator \( \hat{w} \) derived from the training data, we evaluate the test error using the standard method:
\begin{small}
\begin{align}
E_{test}(w) \defeq \mathbb{E}\left[(x_{test}^T w - y_{test})^2 \right] - \sigma^2 = \E[\|w - w^*\|_{\Sigma}^2],\label{eq:test_error}
\end{align}
\end{small}
where the expectation is computed with respect to the training data, while the test pair \( (x_{\text{test}}, y_{\text{test}}) \) is sampled independently from \( P_{\Sigma, w^*, \sigma^2} \) of the training set.
\subsection{Test Error Under Data Editing}
Our goal is to derive an analytical expression for the test error of the \(n\)-th model in the data editing setting. As indicated by the test error in Eq.~\ref{eq:test_error}, this process involves two main steps: (1) establishing the relationship between the fitted linear parameters \(w_n\) and the true parameters \(w^*\), and (2) simplifying the test error expression. We begin by formulating the relationship between \(w_n\) and \(w^*\).
Proofs are detailed in Appendix~\ref{sec:proof}.
\begin{theorem}
In the data editing setting, $\forall n \geq 1$, the fitted linear parameters $\hat{w}_{n+1}$ can be derived as:
\begin{small}
\begin{align}
\hat{w}_{n+1} 
& = w^* + (X^\top X)^{-1} X^\top \left( E_1 + \sum_{i=1}^{n} M_i E_{i+1} \right)
\end{align}
\end{small}
where, $w^*$ is the true parameter, $X$ is the original design matrix, $E_i$ is the extra noise added at the $i$'th iteration, and $M_i$ is an idempotent diagonal matrix, defining the editing operation.
\label{theorem:ridgeless_fitted_linear_parameters}
\end{theorem}

\begin{table*}[t]
\centering
\caption{General performance of the pre-trained base models. PT indicates we pre-train OLMo-1B from scratch.}
\resizebox{0.8\linewidth}{!}{
\begin{tabular}{l|cccccccc|c}
\toprule
& PIQA & BoolQ & OBQA & ARC-c & ARC-e & HellaSwag & SIQA & Winogrande & Avg \\
\midrule
OLMo-1B (PT) & 53.97 & 38.26 & 12.20 & 17.23 & 28.36 & 26.02 & 34.80 & 51.14 & \colorthree{32.75} \\
$\Delta$ ToEdit & 54.13 & 38.65 & 12.80 & 18.43 & 27.48 & 25.94 & 34.95 & 52.49 & \colorfive{\textbf{33.11}} \\
\bottomrule
\end{tabular}}
\label{tab:pt}
\vspace{-1.5em}
\end{table*}
\begin{table*}[th!]
  \footnotesize
  \centering
  \caption{Performance on domain-specific tasks for continual pre-training models. CPT indicates continual pre-training. $\Delta$ denotes training with our edited data. Our method demonstrates consistent improvements across three domains on both OLMo-1B and Llama-3-8B.}
  \resizebox{0.7\linewidth}{!}{
  \begin{tabular}{l|ccccc|c}
      \toprule
      \rowcolor[rgb]{0.93,0.93,0.93}\multicolumn{7}{c}{\textit{Biomedicine}}
      \\
      \midrule
      Models & MQP & ChemProt & PubMedQA  & RCT & USMLE  & Avg \\
      \midrule
      OLMo-1B & 52.59 & 17.2 & 51.40  & 32.70 & 28.90  & \colorone{36.63} \\
      CPT   & 52.29 & 21.00 & 58.50  & 34.90  & 27.49  & \colorthree{38.83} \\
      $\Delta$ ToEdit  & 54.59 & 22.40 & 65.00  & 34.50  & 27.96  & \colorfive{\textbf{40.89}}
      \\
      \midrule
      Llama-3-8B & 66.80 & 28.59 & 60.8 & 73.85 & 40.61 & \colorone{54.13} \\
      CPT  & 72.29 & 29.4 & 69.1  & 72.65  & 36.76  & \colorthree{56.04} \\
      $\Delta$ ToEdit  & 76.39 & 30.2 & 65.3  & 73.30  & 37.23  & \colorfive{\textbf{56.48}}
      \\
      \midrule
      \rowcolor[rgb]{0.93,0.93,0.93}\multicolumn{7}{c}{\textit{Finance}}\\
      \midrule
      Models & HeadLine & FPB & FiQA-SA & ConvFinQA & NER & Avg \\
      \midrule
      OLMo-1B  & 69.00 & 47.03 & 48.05 & 4.83 & 62.19 & \colorone{46.22} \\
      CPT & 70.31 & 49.78 & 40.36 & 18.72 & 60.44 & \colorthree{47.92} \\
      $\Delta$ ToEdit  & 71.77 & 51.39 & 46.06 & 18.85 & 62.97 & \colorfive{\textbf{50.21}} \\
      \midrule
        Llama-3-8B & 81.28 & 63.58 & 81.60 & 52.88 & 72.53 & \colorone{70.37} \\
      CPT  & 85.68 & 54.22 & 81.88  & 67.78  & 67.43  & \colorthree{71.40} \\
      $\Delta$ ToEdit  & 83.83 & 61.61 & 80.82  & 67.31 & 67.62  & \colorfive{\textbf{72.24}}
      \\
      \midrule
      \rowcolor[rgb]{0.93,0.93,0.93}
      \multicolumn{7}{c}{\textit{Math}}
      \\
      \midrule
      Models & ARC-c & GPQA & GSM8K & MATH & MMLU & Avg \\
      \midrule
      OLMo-1B & 28.67 & 24.23 & 1.67 & 0.00 &  26.56 & \colorone{16.23} \\
      CPT & 28.41 & 24.03 & 1.52 & 0.10 &  27.23 & \colorthree{16.26} \\
      $\Delta$ ToEdit  & 28.92 & 28.12 & 2.20 & 0.10  & 23.63 & \colorfive{\textbf{16.59}} \\
      \bottomrule
  \end{tabular}}
  \label{tab:cpt}
\end{table*}

\begin{theorem}
Consider an $n+1$ fold data editing process with $T \geq d + 2$ samples per iteration and isotropic features ($\Sigma \defeq I_d$), the test error for the ridgeless linear model $\hat{w}_n$ learned on the edited data up to iteration $n+1$, is bounded by:
\begin{small}
\begin{align}
E_{test}(\hat{w}_{n+1}) \leq \frac{2\sigma^2d}{T - d - 1}\label{eq:relaxed_test_err} 
\end{align}
\end{small}
Furthermore, assuming the editing operation satisfies $||M_i|| = ||M_{i-1}|| \eta $ with $\eta \in (0, 1)$, the test error can be further bounded by:
\begin{small}
\begin{align}
E_{test}(\hat{w}_{n+1}) 
&\leq \frac{\sigma^2 d}{T - d - 1} \notag \\
&\quad + \sigma^2 \sqrt{\mathbb{E}\left[\text{tr}\left((X^\top X)^{-2}\right)\right]} \cdot \notag \frac{\sqrt{\mathbb{E}\left[\text{tr}(M_1)\right]}}{1 - \eta}.
\end{align}
\end{small}
\end{theorem}\label{theorem:upper_bound}
We provide supporting evidence for the assumption in \S~\ref{sec:Gradual_Decline_in_Editing}.  
Recalling model collapse~\citep{dohmatob2024model}: training iteratively on synthetic data leads to an accumulation of error over iterations, as shown in the following equation:
\begin{small}
\begin{align}
E_{\text{test}}^{\text{collapse}}(\hat{w}_n) = \frac{\sigma^2 d}{T - d - 1} \times n \label{eq:model_collapse}
\end{align}
\end{small}
Comparing Eq.~\ref{eq:relaxed_test_err} with Eq.~\ref{eq:model_collapse}, the test error under data editing is bounded by a fixed value, preventing continuous error accumulation and thus avoiding model collapse. Based on the theoretical derivations and statistical analysis of synthetic data (\S~\ref{sec:non_iterative_model_collapse}), the underlying reason is that our approach retains the coverage of the initial distribution. We move away from pure data synthesis toward token-level editing, which allows us to obtain better data while avoiding model collapse. Moreover, remarkable previous studies~\citep{dohmatob2024tale,gerstgrasser2024model} pointed out similar conclusions. They indicated mixing real data with synthetic data will break model collapse and provide an upper bound under data accumulation. Different from their work, our data editing aims to yield better data, enabling synthetic data to perform well both in theory and practice.

\begin{table}[t!]
\vspace{-1em}
\centering
\caption{Performance of the SFT tasks. We fine-tune LLaMA-3-8B using instruction tuning and code reasoning tasks, comparing performance with the edited version produced by our method. HS and WG are short for HellaSwag and Winogrande respectively.
}
\resizebox{\linewidth}{!}{
\begin{tabular}{l|l|ccccc|c}
\toprule
& Models & PIQA & BoolQ & HS & SIQA & WG & Avg \\
\midrule
\rowcolor[rgb]{0.93,0.93,0.93}\multicolumn{8}{c}{\textit{Instruction Tuning}} \\
\midrule
{\textit{Natural}} & Llama-3 & 79.82 & 87.06  & 58.32 & 46.83 & 74.66 & \colorthree{69.34} \\
{\textit{Ins.}} &$\Delta$ ToEdit & 80.58 & 87.80 & 58.27 & 46.93 & 74.90 & \colorfive{\textbf{69.70}}
\\
\midrule
\multirow{2}{*}{\textit{CoT}} & Llama-3 & 79.87 & 81.28 & 59.72& 49.69 & 74.51 & \colorthree{69.01}   \\ 
& $\Delta$ ToEdit & 80.25 & 81.16 & 59.74 & 50.56 & 74.59 & \colorfive{\textbf{69.26}} \\ 
\midrule
\multirow{2}{*}{\textit{FLANv2}} & Llama-3 & 80.79 & 84.04 & 59.98 & 51.43 & 74.66 & \colorthree{70.18} \\ 
& $\Delta$ ToEdit & 80.69 & 85.20 & 59.99 & 52.00 & 75.37& \colorfive{\textbf{70.65}} \\ 
\midrule
{\textit{Open}} & Llama-3 & 79.65 & 83.18 & 60.51 & 48.52 & 74.11 & \colorthree{69.19} \\ 
{\textit{Assist.}} & $\Delta$ ToEdit & 79.98 & 83.91 & 60.34 & 48.31 & 74.66 & \colorfive{\textbf{69.44}}\\
\bottomrule
\end{tabular}}

\resizebox{\linewidth}{!}{
\begin{tabular}{l|l|cccc|c}
\toprule
& Models & ARC-c & GPQA  & GSM8K & MMLU & Avg \\
\midrule
\rowcolor[rgb]{0.93,0.93,0.93}\multicolumn{7}{c}{\textit{Code Reasoning}} \\
\midrule
{\textit{OSS-}} & Llama-3 & 51.28 & 27.46 & 49.58 & 62.14 & \colorthree{45.76 } \\ 
{\textit{Inst.}}& $\Delta$ ToEdit & 51.79 & 28.79 & 49.36 & 62.04 & \colorfive{\textbf{46.13}} \\
\midrule
{\textit{Evol-}}& Llama-3 & 52.90 & 27.90 & 50.87 & 62.40 & \colorthree{46.62} \\ 
{\textit{Ins.}}& $\Delta$ ToEdit & 52.22 & 29.69 & 50.87 & 62.60 & \colorfive{\textbf{46.92}} \\ 
\bottomrule
\end{tabular}}
\label{tab:sft}
\end{table}

\section{Experiments}~\label{sec:experiments}We validate our method in three language model training stages: pre-training from scratch (PT), continual pre-training (CPT), and supervised fine-tuning (SFT). 
\subsection{Implementation}
We use the Llama-3-8B~\citep{llama3modelcard} as a prior distribution to estimate the token distribution in each text sample. The modification probability is set to $p = 0.99$. This means that we resample tokens in positions where the probability exceeds $p$, and the resampling is based on the conditional probability given the preceding context. The entire process requires only a single forward pass, without auto-regressive generation. We integrate the fast inference engine vLLM~\citep{kwon2023efficient}, allowing the entire data editing process to be completed on a single 4090 GPU. 
We use top-k as the sampling strategy with $k = 8$. We also incorporate top-p sampling and rejection sampling in our ablation studies.
\vspace{-1em}
\subsection{Datasets and Models} 
We provide an overview of our experimental setup, more details are in Appendix~\ref{sec:experiment_setting}. 
\textbf{For pre-training}, we pre-train the 1B OLMo model~\citep{OLMo} from scratch using Dolma-sampled V6 (6B tokens) and evaluate on 8 general tasks. \textbf{For continual pre-training}, we follow~\cite{Cheng2024InstructionPL,cheng2024adapting,cheng2024domain} to continual pre-train OLMo-1B~\citep{OLMo} and Llama-3-8B~\citep{llama3modelcard} on corpora of Biomedicine, Finance, and Math, evaluating on 5 downstream tasks per domain. \textbf{For supervised fine-tuning}, we fine-tune Llama-3-8B on instruction tuning and code reasoning tasks, evaluating on 9 downstream tasks.

\subsection{Main Results}
Table~\ref{tab:pt},~\ref{tab:cpt}, and~\ref{tab:sft} respectively demonstrate the effectiveness of our method in pre-training from scratch, continual pre-training and fine-tuning tasks. Across these three stages of language model training, our method consistently enhances the model performance on downstream tasks without increasing the data size. This consistency is validated across two models. This indicates that our method unlocks the potential of existing data, demonstrating that semi-synthetic data is a viable path to improving model performance. A further numerical analysis is provided in \S~\ref{sec:More_Discussion_of_Main_Results}.
\begin{table}[t!]
\centering
\caption{
Ablations on resampled token condition ($p$) in biomedicine domain.}
\label{tab:ablation_resampled_p}
\resizebox{\linewidth}{!}{
\begin{tabular}{l|ccccc|c}
\toprule
& PubMedQA & MQP   & RCT   & USMLE & ChemProt & Avg    \\ 
\midrule
$p \geq 0.99$  & 64.50     & 55.73 & 30.95 & 27.65 & 14.60     & 38.69 \\ 
$p \geq 0.999$ & 63.60     & 55.40  & 29.09 & 28.12 & 16.20     & 38.48 \\ 
$p \leq 0.1$   & 62.40     & 51.47 & 25.60  & 29.14 & 10.00     & 35.72 \\ 
$p \leq 0.01$  & 65.40     & 54.91 & 28.19 & 27.80 & 11.00     & 37.46  \\ 
\bottomrule
\end{tabular}} 
\end{table}
\begin{table}[t!]
\centering
\caption{
Token distribution across different probability intervals in the biomedicine domain dataset.
}\label{tab:token_distribution}
\resizebox{\linewidth}{!}{
\begin{tabular}{ccc|ccc}
\toprule
{Interval} & {Percent} & {\# Tokens} & {Interval} & {Percent} & {\# Tokens} \\ 
\midrule
$[0.0, 0.1)$ & 34.7\% & 389M & $[0.5, 0.6)$ & 3.6\% & 40M \\
$[0.1, 0.2)$ & 8.1\% & 91M & $[0.6, 0.7)$ & 3.7\% & 41M \\
$[0.2, 0.3)$ & 5.4\% & 60M & $[0.7, 0.8)$ & 4.0\% & 44M \\
$[0.3, 0.4)$ & 4.4\% & 49M & $[0.8, 0.9)$ & 5.2\% & 58M \\
$[0.4, 0.5)$ & 3.8\% & 43M & $[0.9, 1.0)$ & 27.1\% & 303M \\
\bottomrule
\end{tabular}}
\end{table}
\begin{table}[t]
\centering
\caption{Ablations on sampling strategy.}\label{tab:sampling_strategies}
\resizebox{0.9\linewidth}{!}{
\begin{tabular}{l|ccc}
\toprule
Strategy & PubMedQA & MedMCQA & MedQA \\ 
\midrule
Top-k             & 64.50     & 26.13   & 24.82          \\ 
Top-p             & 63.80     & 27.11   & 25.61          \\ 
Rejection Sampling   & 64.50     & 28.90   & 28.20          \\ 
\bottomrule
\end{tabular}
} 
\end{table}
\begin{table}[t]
\centering
\caption{Ablations on sampling size $k$ for top-k.}\label{tab:ablation_sampling_size}
\resizebox{0.9\linewidth}{!}{
\begin{tabular}{l|ccc}
\toprule
Sampling Size ($k$) & PubMedQA & MedMCQA & MedQA \\ 
\midrule
$k=8$   & 64.50     & 26.13   & 24.82          \\ 
$k=64$  & 63.80     & 28.14   & 27.34          \\ 
\bottomrule
\end{tabular}} 
\end{table}

\subsection{Ablation Studies}\label{sec:ablation}


We conduct experiments on hyper-parameter $p$, including: (1) ablation studies on different values, (2) token percentage statistics, (3) comparisons of sampling strategies, and (4) an ablation study on sampling size.

Table~\ref{tab:ablation_resampled_p} shows the impact of different $p$ values on the model performance, with fluctuations observed across various settings. Table~\ref{tab:token_distribution} presents the distribution percentages across different probability value ranges. As mentioned above, we need to refine the data while preserving mainly source distribution. As shown in Figure~\ref{fig:qwen2_probs}, a larger $p$ indicates fewer tokens will be resampled, while a smaller $p$ results in more tokens being resampled. To balance model performance and data distribution preservation, we set $p=0.99$ as threshold for our experiments. Table~\ref{tab:sampling_strategies} presents the results of different sampling strategies. Specifically, to control variables, we set $k=8$ for top-k sampling and $p=0.99$ for top-p sampling. We use rejection sampling implementation in~\cite{Liu2023StatisticalRS}. The results of reject sampling, top-p, and top-k are comparable. However, top-p involves a dynamic sampling range, and reject sampling requires multiple rounds of computation, leading to increased overhead. Considering computational efficiency, we choose top-k for sampling. Table~\ref{tab:ablation_sampling_size} shows the ablation study on sampling size of top-k.  The performance gain from increasing $k$ is relatively small. Therefore, we set $k=8$ in our experiments. And a detailed case for token editing is provided in Table~\ref{tab:case_study}.




\section{Conclusion}

With the growing prevalence of generative AI models, when training next-generation AI models, it will be inevitable to use a mixture of synthetic data and human-produced data. Therefore, we focus on two key questions: (1) What is the impact of synthetic data on language model pre-training, and what are the underlying causes? (2) How can we prevent model collapse and synthesize high-quality data? We found that synthetic data can impair the effectiveness of pre-training when mixed with human-produced data, leading to non-iterative model collapse. Statistical analysis reveals that synthetic data suffers from significant distribution gaps and overly concentrated n-gram features. We propose token-level editing instead of relying purely on synthetic data. Specifically, we perform token resampling guided by a trained prior. Theoretically, our method can prevent model collapse. Our approach demonstrates improvements over the source data across pre-training, continual pre-training, and supervised fine-tuning.

\section*{Impact Statement}
This paper presents work whose goal is to advance the field of Machine Learning. There are many potential societal consequences of our work, none which we feel must be specifically highlighted here.

\section*{Acknowledgement}
This work is sponsored by the National Key Research and Development Program of China (No. 2023ZD0121402). X.Z., H.L. and Z.Z. are supported by the National Natural Science Foundation of China (62376031).


\bibliography{example_paper}

\begin{thebibliography}{70}
\providecommand{\natexlab}[1]{#1}
\providecommand{\url}[1]{\texttt{#1}}
\expandafter\ifx\csname urlstyle\endcsname\relax
  \providecommand{\doi}[1]{doi: #1}\else
  \providecommand{\doi}{doi: \begingroup \urlstyle{rm}\Url}\fi

\bibitem[qwe(2024)]{qwen2}
Qwen2 technical report.
\newblock 2024.

\bibitem[Achiam et~al.(2023)Achiam, Adler, Agarwal, Ahmad, Akkaya, Aleman, Almeida, Altenschmidt, Altman, Anadkat, et~al.]{achiam2023gpt}
Achiam, J., Adler, S., Agarwal, S., Ahmad, L., Akkaya, I., Aleman, F.~L., Almeida, D., Altenschmidt, J., Altman, S., Anadkat, S., et~al.
\newblock Gpt-4 technical report.
\newblock \emph{arXiv preprint arXiv:2303.08774}, 2023.

\bibitem[AI@Meta(2024)]{llama3modelcard}
AI@Meta.
\newblock Llama 3 model card.
\newblock 2024.
\newblock URL \url{https://github.com/meta-llama/llama3/blob/main/MODEL_CARD.md}.

\bibitem[Albalak et~al.(2024)Albalak, Elazar, Xie, Longpre, Lambert, Wang, Muennighoff, Hou, Pan, Jeong, et~al.]{albalak2024survey}
Albalak, A., Elazar, Y., Xie, S.~M., Longpre, S., Lambert, N., Wang, X., Muennighoff, N., Hou, B., Pan, L., Jeong, H., et~al.
\newblock A survey on data selection for language models.
\newblock \emph{arXiv preprint arXiv:2402.16827}, 2024.

\bibitem[Alemohammad et~al.(2023)Alemohammad, Casco-Rodriguez, Luzi, Humayun, Babaei, LeJeune, Siahkoohi, and Baraniuk]{alemohammad2023self}
Alemohammad, S., Casco-Rodriguez, J., Luzi, L., Humayun, A.~I., Babaei, H., LeJeune, D., Siahkoohi, A., and Baraniuk, R.~G.
\newblock Self-consuming generative models go mad.
\newblock \emph{arXiv preprint arXiv:2307.01850}, 4:\penalty0 14, 2023.

\bibitem[Ankner et~al.(2024)Ankner, Blakeney, Sreenivasan, Marion, Leavitt, and Paul]{ankner2024perplexed}
Ankner, Z., Blakeney, C., Sreenivasan, K., Marion, M., Leavitt, M.~L., and Paul, M.
\newblock Perplexed by perplexity: Perplexity-based data pruning with small reference models.
\newblock \emph{arXiv preprint arXiv:2405.20541}, 2024.

\bibitem[Bai et~al.(2023)Bai, Zhang, Tao, Wu, Wang, and Xu]{Bai_Zhang_Tao_Wu_Wang_Xu_2023}
Bai, F., Zhang, H., Tao, T., Wu, Z., Wang, Y., and Xu, B.
\newblock Picor: Multi-task deep reinforcement learning with policy correction.
\newblock \emph{Proceedings of the AAAI Conference on Artificial Intelligence}, 37\penalty0 (6):\penalty0 6728--6736, Jun. 2023.

\bibitem[Bai et~al.(2024)Bai, Wang, Zhang, Chen, Xu, Wen, and Yang]{bai2024efficient}
Bai, F., Wang, M., Zhang, Z., Chen, B., Xu, Y., Wen, Y., and Yang, Y.
\newblock Efficient model-agnostic alignment via bayesian persuasion.
\newblock \emph{arXiv preprint arXiv:2405.18718}, 2024.

\bibitem[Bai et~al.(2025)Bai, Liu, Du, Wen, and Yang]{bai2025rat}
Bai, F., Liu, R., Du, Y., Wen, Y., and Yang, Y.
\newblock Rat: Adversarial attacks on deep reinforcement agents for targeted behaviors.
\newblock In \emph{Proceedings of the AAAI Conference on Artificial Intelligence}, volume~39, pp.\  15453--15461, 2025.

\bibitem[Bellagente et~al.(2024)Bellagente, Tow, Mahan, Phung, Zhuravinskyi, Adithyan, Baicoianu, Brooks, Cooper, Datta, et~al.]{bellagente2024stable}
Bellagente, M., Tow, J., Mahan, D., Phung, D., Zhuravinskyi, M., Adithyan, R., Baicoianu, J., Brooks, B., Cooper, N., Datta, A., et~al.
\newblock Stable lm 2 1.6 b technical report.
\newblock \emph{arXiv preprint arXiv:2402.17834}, 2024.

\bibitem[Ben~Allal et~al.(2024)Ben~Allal, Lozhkov, Penedo, Wolf, and von Werra]{benallal2024cosmopedia}
Ben~Allal, L., Lozhkov, A., Penedo, G., Wolf, T., and von Werra, L.
\newblock Cosmopedia, 2024.
\newblock URL \url{https://huggingface.co/datasets/HuggingFaceTB/cosmopedia}.

\bibitem[Bertrand et~al.(2023)Bertrand, Bose, Duplessis, Jiralerspong, and Gidel]{bertrand2023stability}
Bertrand, Q., Bose, A.~J., Duplessis, A., Jiralerspong, M., and Gidel, G.
\newblock On the stability of iterative retraining of generative models on their own data.
\newblock \emph{arXiv preprint arXiv:2310.00429}, 2023.

\bibitem[Briesch et~al.(2023)Briesch, Sobania, and Rothlauf]{briesch2023large}
Briesch, M., Sobania, D., and Rothlauf, F.
\newblock Large language models suffer from their own output: An analysis of the self-consuming training loop.
\newblock \emph{arXiv preprint arXiv:2311.16822}, 2023.

\bibitem[Cheng et~al.(2024{\natexlab{a}})Cheng, Gu, Huang, Bi, Huang, and Wei]{Cheng2024InstructionPL}
Cheng, D., Gu, Y., Huang, S., Bi, J., Huang, M., and Wei, F.
\newblock Instruction pre-training: Language models are supervised multitask learners.
\newblock In \emph{Conference on Empirical Methods in Natural Language Processing}, 2024{\natexlab{a}}.
\newblock URL \url{https://api.semanticscholar.org/CorpusID:270620509}.

\bibitem[Cheng et~al.(2024{\natexlab{b}})Cheng, Huang, and Wei]{cheng2024adapting}
Cheng, D., Huang, S., and Wei, F.
\newblock Adapting large language models via reading comprehension.
\newblock In \emph{The Twelfth International Conference on Learning Representations}, 2024{\natexlab{b}}.
\newblock URL \url{https://openreview.net/forum?id=y886UXPEZ0}.

\bibitem[Cheng et~al.(2024{\natexlab{c}})Cheng, Huang, Zhu, Zhang, Zhao, Luan, Dai, and Zhang]{cheng2024domain}
Cheng, D., Huang, S., Zhu, Z., Zhang, X., Zhao, W.~X., Luan, Z., Dai, B., and Zhang, Z.
\newblock On domain-specific post-training for multimodal large language models.
\newblock \emph{arXiv preprint arXiv:2411.19930}, 2024{\natexlab{c}}.

\bibitem[Dohmatob et~al.(2024{\natexlab{a}})Dohmatob, Feng, and Kempe]{dohmatob2024model}
Dohmatob, E., Feng, Y., and Kempe, J.
\newblock Model collapse demystified: The case of regression.
\newblock \emph{arXiv preprint arXiv:2402.07712}, 2024{\natexlab{a}}.

\bibitem[Dohmatob et~al.(2024{\natexlab{b}})Dohmatob, Feng, Subramonian, and Kempe]{dohmatob2024strong}
Dohmatob, E., Feng, Y., Subramonian, A., and Kempe, J.
\newblock Strong model collapse.
\newblock \emph{arXiv preprint arXiv:2410.04840}, 2024{\natexlab{b}}.

\bibitem[Dohmatob et~al.(2024{\natexlab{c}})Dohmatob, Feng, Yang, Charton, and Kempe]{dohmatob2024tale}
Dohmatob, E., Feng, Y., Yang, P., Charton, F., and Kempe, J.
\newblock A tale of tails: Model collapse as a change of scaling laws.
\newblock \emph{arXiv preprint arXiv:2402.07043}, 2024{\natexlab{c}}.

\bibitem[Eldan \& Li(2023)Eldan and Li]{eldan2023tinystories}
Eldan, R. and Li, Y.
\newblock Tinystories: How small can language models be and still speak coherent english?
\newblock \emph{arXiv preprint arXiv:2305.07759}, 2023.

\bibitem[Feng et~al.(2024)Feng, Dohmatob, Yang, Charton, Kempe, and Meta]{feng2024beyond}
Feng, Y., Dohmatob, E., Yang, P., Charton, F., Kempe, J., and Meta, F.
\newblock Beyond model collapse: Scaling up with syn-thesized data requires verification.
\newblock \emph{arXiv preprint arXiv:2406.07515}, 2024.

\bibitem[Ferbach et~al.(2024)Ferbach, Bertrand, Bose, and Gidel]{ferbach2024self}
Ferbach, D., Bertrand, Q., Bose, A.~J., and Gidel, G.
\newblock Self-consuming generative models with curated data provably optimize human preferences.
\newblock \emph{arXiv preprint arXiv:2407.09499}, 2024.

\bibitem[Gao et~al.(2020{\natexlab{a}})Gao, Biderman, Black, Golding, Hoppe, Foster, Phang, He, Thite, Nabeshima, Presser, and Leahy]{Gao2020ThePA}
Gao, L., Biderman, S., Black, S., Golding, L., Hoppe, T., Foster, C., Phang, J., He, H., Thite, A., Nabeshima, N., Presser, S., and Leahy, C.
\newblock The pile: An 800gb dataset of diverse text for language modeling.
\newblock \emph{ArXiv}, abs/2101.00027, 2020{\natexlab{a}}.
\newblock URL \url{https://api.semanticscholar.org/CorpusID:230435736}.

\bibitem[Gao et~al.(2020{\natexlab{b}})Gao, Biderman, Black, Golding, Hoppe, Foster, Phang, He, Thite, Nabeshima, et~al.]{gao2020pile}
Gao, L., Biderman, S., Black, S., Golding, L., Hoppe, T., Foster, C., Phang, J., He, H., Thite, A., Nabeshima, N., et~al.
\newblock The pile: An 800gb dataset of diverse text for language modeling.
\newblock \emph{arXiv preprint arXiv:2101.00027}, 2020{\natexlab{b}}.

\bibitem[Gao et~al.(2024)Gao, Tow, Abbasi, Biderman, Black, DiPofi, Foster, Golding, Hsu, Le~Noac'h, Li, McDonell, Muennighoff, Ociepa, Phang, Reynolds, Schoelkopf, Skowron, Sutawika, Tang, Thite, Wang, Wang, and Zou]{eval-harness}
Gao, L., Tow, J., Abbasi, B., Biderman, S., Black, S., DiPofi, A., Foster, C., Golding, L., Hsu, J., Le~Noac'h, A., Li, H., McDonell, K., Muennighoff, N., Ociepa, C., Phang, J., Reynolds, L., Schoelkopf, H., Skowron, A., Sutawika, L., Tang, E., Thite, A., Wang, B., Wang, K., and Zou, A.
\newblock A framework for few-shot language model evaluation, 07 2024.
\newblock URL \url{https://zenodo.org/records/12608602}.

\bibitem[Gerstgrasser et~al.(2024)Gerstgrasser, Schaeffer, Dey, Rafailov, Sleight, Hughes, Korbak, Agrawal, Pai, Gromov, et~al.]{gerstgrasser2024model}
Gerstgrasser, M., Schaeffer, R., Dey, A., Rafailov, R., Sleight, H., Hughes, J., Korbak, T., Agrawal, R., Pai, D., Gromov, A., et~al.
\newblock Is model collapse inevitable? breaking the curse of recursion by accumulating real and synthetic data.
\newblock \emph{arXiv preprint arXiv:2404.01413}, 2024.

\bibitem[Groeneveld et~al.(2024)Groeneveld, Beltagy, Walsh, Bhagia, Kinney, Tafjord, Jha, Ivison, Magnusson, Wang, Arora, Atkinson, Authur, Chandu, Cohan, Dumas, Elazar, Gu, Hessel, Khot, Merrill, Morrison, Muennighoff, Naik, Nam, Peters, Pyatkin, Ravichander, Schwenk, Shah, Smith, Strubell, Subramani, Wortsman, Dasigi, Lambert, Richardson, Zettlemoyer, Dodge, Lo, Soldaini, Smith, and Hajishirzi]{OLMo}
Groeneveld, D., Beltagy, I., Walsh, P., Bhagia, A., Kinney, R., Tafjord, O., Jha, A., Ivison, H., Magnusson, I., Wang, Y., Arora, S., Atkinson, D., Authur, R., Chandu, K.~R., Cohan, A., Dumas, J., Elazar, Y., Gu, Y., Hessel, J., Khot, T., Merrill, W., Morrison, J.~D., Muennighoff, N., Naik, A., Nam, C., Peters, M.~E., Pyatkin, V., Ravichander, A., Schwenk, D., Shah, S., Smith, W., Strubell, E., Subramani, N., Wortsman, M., Dasigi, P., Lambert, N., Richardson, K., Zettlemoyer, L., Dodge, J., Lo, K., Soldaini, L., Smith, N.~A., and Hajishirzi, H.
\newblock Olmo: Accelerating the science of language models.
\newblock \emph{arXiv preprint}, 2024.
\newblock URL \url{https://api.semanticscholar.org/CorpusID:267365485}.

\bibitem[Gulcehre et~al.(2023)Gulcehre, Paine, Srinivasan, Konyushkova, Weerts, Sharma, Siddhant, Ahern, Wang, Gu, et~al.]{gulcehre2023reinforced}
Gulcehre, C., Paine, T.~L., Srinivasan, S., Konyushkova, K., Weerts, L., Sharma, A., Siddhant, A., Ahern, A., Wang, M., Gu, C., et~al.
\newblock Reinforced self-training (rest) for language modeling.
\newblock \emph{arXiv preprint arXiv:2308.08998}, 2023.

\bibitem[Gunasekar et~al.(2023)Gunasekar, Zhang, Aneja, Mendes, Del~Giorno, Gopi, Javaheripi, Kauffmann, de~Rosa, Saarikivi, et~al.]{gunasekar2023textbooks}
Gunasekar, S., Zhang, Y., Aneja, J., Mendes, C. C.~T., Del~Giorno, A., Gopi, S., Javaheripi, M., Kauffmann, P., de~Rosa, G., Saarikivi, O., et~al.
\newblock Textbooks are all you need.
\newblock \emph{arXiv preprint arXiv:2306.11644}, 2023.

\bibitem[Hu et~al.(2021)Hu, Shen, Wallis, Allen-Zhu, Li, Wang, Wang, and Chen]{hu2021lora}
Hu, E.~J., Shen, Y., Wallis, P., Allen-Zhu, Z., Li, Y., Wang, S., Wang, L., and Chen, W.
\newblock Lora: Low-rank adaptation of large language models.
\newblock \emph{arXiv preprint arXiv:2106.09685}, 2021.

\bibitem[Jia et~al.(2024)Jia, Yang, Li, Zhang, and Yan]{jia2024bench2drive}
Jia, X., Yang, Z., Li, Q., Zhang, Z., and Yan, J.
\newblock Bench2drive: Towards multi-ability benchmarking of closed-loop end-to-end autonomous driving.
\newblock \emph{arXiv preprint arXiv:2406.03877}, 2024.

\bibitem[Jiang et~al.(2024)Jiang, Sablayrolles, Roux, Mensch, Savary, Bamford, Chaplot, Casas, Hanna, Bressand, et~al.]{jiang2024mixtral}
Jiang, A.~Q., Sablayrolles, A., Roux, A., Mensch, A., Savary, B., Bamford, C., Chaplot, D.~S., Casas, D. d.~l., Hanna, E.~B., Bressand, F., et~al.
\newblock Mixtral of experts.
\newblock \emph{arXiv preprint arXiv:2401.04088}, 2024.

\bibitem[Kazdan et~al.(2025)Kazdan, Schaeffer, Dey, Gerstgrasser, Rafailov, Donoho, and Koyejo]{kazdan2025collapse}
Kazdan, J., Schaeffer, R., Dey, A., Gerstgrasser, M., Rafailov, R., Donoho, D.~L., and Koyejo, S.
\newblock Collapse or thrive? perils and promises of synthetic data in a self-generating world, 2025.
\newblock URL \url{https://openreview.net/forum?id=Xr5iINA3zU}.

\bibitem[Kopf et~al.(2023)Kopf, Kilcher, von Rutte, Anagnostidis, Tam, Stevens, Barhoum, Duc, Stanley, Nagyfi, Shahul, Suri, Glushkov, Dantuluri, Maguire, Schuhmann, Nguyen, and Mattick]{Kopf2023OpenAssistantC}
Kopf, A., Kilcher, Y., von Rutte, D., Anagnostidis, S., Tam, Z.~R., Stevens, K., Barhoum, A., Duc, N.~M., Stanley, O., Nagyfi, R., Shahul, E., Suri, S., Glushkov, D., Dantuluri, A., Maguire, A., Schuhmann, C., Nguyen, H., and Mattick, A.
\newblock Openassistant conversations - democratizing large language model alignment.
\newblock \emph{ArXiv}, abs/2304.07327, 2023.
\newblock URL \url{https://api.semanticscholar.org/CorpusID:258179434}.

\bibitem[Kwon et~al.(2023)Kwon, Li, Zhuang, Sheng, Zheng, Yu, Gonzalez, Zhang, and Stoica]{kwon2023efficient}
Kwon, W., Li, Z., Zhuang, S., Sheng, Y., Zheng, L., Yu, C.~H., Gonzalez, J.~E., Zhang, H., and Stoica, I.
\newblock Efficient memory management for large language model serving with pagedattention.
\newblock In \emph{Proceedings of the ACM SIGOPS 29th Symposium on Operating Systems Principles}, 2023.

\bibitem[Li et~al.(2023)Li, Lyu, Ma, Wang, Yang, Li, and Li]{li2023normalization}
Li, L., Lyu, J., Ma, G., Wang, Z., Yang, Z., Li, X., and Li, Z.
\newblock Normalization enhances generalization in visual reinforcement learning.
\newblock \emph{arXiv preprint arXiv:2306.00656}, 2023.

\bibitem[Li et~al.(2024)Li, Jia, Wang, and Yan]{li2024think2drive}
Li, Q., Jia, X., Wang, S., and Yan, J.
\newblock Think2drive: Efficient reinforcement learning by thinking in latent world model for quasi-realistic autonomous driving (in carla-v2).
\newblock \emph{arXiv preprint arXiv:2402.16720}, 2024.

\bibitem[Li et~al.(2020)Li, Yang, and Wu]{li2020face}
Li, X., Yang, Z., and Wu, H.
\newblock Face detection based on receptive field enhanced multi-task cascaded convolutional neural networks.
\newblock \emph{IEEE access}, 8:\penalty0 174922--174930, 2020.

\bibitem[Lin et~al.(2024)Lin, Gou, Gong, Liu, Shen, Xu, Lin, Yang, Jiao, Duan, et~al.]{lin2024rho}
Lin, Z., Gou, Z., Gong, Y., Liu, X., Shen, Y., Xu, R., Lin, C., Yang, Y., Jiao, J., Duan, N., et~al.
\newblock Rho-1: Not all tokens are what you need.
\newblock \emph{arXiv preprint arXiv:2404.07965}, 2024.

\bibitem[Liu et~al.(2024)Liu, Wei, Liu, Si, Zhang, Rao, Zheng, Peng, Yang, Zhou, et~al.]{liu2024best}
Liu, R., Wei, J., Liu, F., Si, C., Zhang, Y., Rao, J., Zheng, S., Peng, D., Yang, D., Zhou, D., et~al.
\newblock Best practices and lessons learned on synthetic data for language models.
\newblock \emph{arXiv preprint arXiv:2404.07503}, 2024.

\bibitem[Liu et~al.(2023)Liu, Zhao, Joshi, Khalman, Saleh, Liu, and Liu]{Liu2023StatisticalRS}
Liu, T., Zhao, Y., Joshi, R., Khalman, M., Saleh, M., Liu, P.~J., and Liu, J.
\newblock Statistical rejection sampling improves preference optimization.
\newblock \emph{ArXiv}, abs/2309.06657, 2023.
\newblock URL \url{https://api.semanticscholar.org/CorpusID:261705578}.

\bibitem[Longpre et~al.(2023)Longpre, Hou, Vu, Webson, Chung, Tay, Zhou, Le, Zoph, Wei, et~al.]{longpre2023flan}
Longpre, S., Hou, L., Vu, T., Webson, A., Chung, H.~W., Tay, Y., Zhou, D., Le, Q.~V., Zoph, B., Wei, J., et~al.
\newblock The flan collection: Designing data and methods for effective instruction tuning.
\newblock \emph{arXiv preprint arXiv:2301.13688}, 2023.

\bibitem[Magnusson et~al.(2023)Magnusson, Bhagia, Hofmann, Soldaini, Jha, Tafjord, Schwenk, Walsh, Elazar, Lo, Groeneveld, Beltagy, Hajishirzi, Smith, Richardson, and Dodge]{Magnusson2023PalomaAB}
Magnusson, I., Bhagia, A., Hofmann, V., Soldaini, L., Jha, A., Tafjord, O., Schwenk, D., Walsh, P., Elazar, Y., Lo, K., Groeneveld, D., Beltagy, I., Hajishirzi, H., Smith, N.~A., Richardson, K., and Dodge, J.
\newblock Paloma: A benchmark for evaluating language model fit.
\newblock \emph{ArXiv}, abs/2312.10523, 2023.
\newblock URL \url{https://api.semanticscholar.org/CorpusID:266348815}.

\bibitem[Maini et~al.(2024)Maini, Seto, Bai, Grangier, Zhang, and Jaitly]{Maini2024RephrasingTW}
Maini, P., Seto, S., Bai, R.~H., Grangier, D., Zhang, Y., and Jaitly, N.
\newblock Rephrasing the web: A recipe for compute and data-efficient language modeling.
\newblock In \emph{Annual Meeting of the Association for Computational Linguistics}, 2024.
\newblock URL \url{https://api.semanticscholar.org/CorpusID:267312030}.

\bibitem[Mart{\'\i}nez et~al.(2023)Mart{\'\i}nez, Watson, Reviriego, Hern{\'a}ndez, Juarez, and Sarkar]{martinez2023towards}
Mart{\'\i}nez, G., Watson, L., Reviriego, P., Hern{\'a}ndez, J.~A., Juarez, M., and Sarkar, R.
\newblock Towards understanding the interplay of generative artificial intelligence and the internet.
\newblock In \emph{International Workshop on Epistemic Uncertainty in Artificial Intelligence}, pp.\  59--73. Springer, 2023.

\bibitem[Mindermann et~al.(2022)Mindermann, Brauner, Razzak, Sharma, Kirsch, Xu, H{\"o}ltgen, Gomez, Morisot, Farquhar, et~al.]{mindermann2022prioritized}
Mindermann, S., Brauner, J.~M., Razzak, M.~T., Sharma, M., Kirsch, A., Xu, W., H{\"o}ltgen, B., Gomez, A.~N., Morisot, A., Farquhar, S., et~al.
\newblock Prioritized training on points that are learnable, worth learning, and not yet learnt.
\newblock In \emph{International Conference on Machine Learning}, pp.\  15630--15649. PMLR, 2022.

\bibitem[Mobahi et~al.(2020)Mobahi, Farajtabar, and Bartlett]{mobahi2020self}
Mobahi, H., Farajtabar, M., and Bartlett, P.
\newblock Self-distillation amplifies regularization in hilbert space.
\newblock \emph{Advances in Neural Information Processing Systems}, 33:\penalty0 3351--3361, 2020.

\bibitem[Niu et~al.(2024)Niu, Pu, Yang, Li, Zhou, Ren, Hu, Li, and Liu]{niu2024lightzero}
Niu, Y., Pu, Y., Yang, Z., Li, X., Zhou, T., Ren, J., Hu, S., Li, H., and Liu, Y.
\newblock Lightzero: A unified benchmark for monte carlo tree search in general sequential decision scenarios.
\newblock \emph{Advances in Neural Information Processing Systems}, 36, 2024.

\bibitem[Pu et~al.(2024)Pu, Niu, Yang, Ren, Li, and Liu]{pu2024unizero}
Pu, Y., Niu, Y., Yang, Z., Ren, J., Li, H., and Liu, Y.
\newblock Unizero: Generalized and efficient planning with scalable latent world models.
\newblock \emph{arXiv preprint arXiv:2406.10667}, 2024.

\bibitem[Radford et~al.(2019)Radford, Wu, Child, Luan, Amodei, and Sutskever]{radford2019language}
Radford, A., Wu, J., Child, R., Luan, D., Amodei, D., and Sutskever, I.
\newblock Language models are unsupervised multitask learners.
\newblock 2019.

\bibitem[Rombach et~al.(2021)Rombach, Blattmann, Lorenz, Esser, and Ommer]{rombach2021highresolution}
Rombach, R., Blattmann, A., Lorenz, D., Esser, P., and Ommer, B.
\newblock High-resolution image synthesis with latent diffusion models, 2021.

\bibitem[Rudin(1976)]{rudin1976principles}
Rudin, W.
\newblock \emph{Principles of Mathematical Analysis}.
\newblock McGraw-Hill, New York, 3rd edition, 1976.

\bibitem[Seddik et~al.(2024)Seddik, Chen, Hayou, Youssef, and Debbah]{seddik2024bad}
Seddik, M. E.~A., Chen, S.-W., Hayou, S., Youssef, P., and Debbah, M.
\newblock How bad is training on synthetic data? a statistical analysis of language model collapse.
\newblock \emph{arXiv preprint arXiv:2404.05090}, 2024.

\bibitem[Shumailov et~al.(2024)Shumailov, Shumaylov, Zhao, Papernot, Anderson, and Gal]{shumailov2024ai}
Shumailov, I., Shumaylov, Z., Zhao, Y., Papernot, N., Anderson, R., and Gal, Y.
\newblock Ai models collapse when trained on recursively generated data.
\newblock \emph{Nature}, 631\penalty0 (8022):\penalty0 755--759, 2024.

\bibitem[Singh et~al.(2023)Singh, Co-Reyes, Agarwal, Anand, Patil, Garcia, Liu, Harrison, Lee, Xu, et~al.]{singh2023beyond}
Singh, A., Co-Reyes, J.~D., Agarwal, R., Anand, A., Patil, P., Garcia, X., Liu, P.~J., Harrison, J., Lee, J., Xu, K., et~al.
\newblock Beyond human data: Scaling self-training for problem-solving with language models.
\newblock \emph{arXiv preprint arXiv:2312.06585}, 2023.

\bibitem[Soldaini et~al.(2024)Soldaini, Kinney, Bhagia, Schwenk, Atkinson, Authur, Bogin, Chandu, Dumas, Elazar, Hofmann, Jha, Kumar, Lucy, Lyu, Lambert, Magnusson, Morrison, Muennighoff, Naik, Nam, Peters, Ravichander, Richardson, Shen, Strubell, Subramani, Tafjord, Walsh, Zettlemoyer, Smith, Hajishirzi, Beltagy, Groeneveld, Dodge, and Lo]{dolma}
Soldaini, L., Kinney, R., Bhagia, A., Schwenk, D., Atkinson, D., Authur, R., Bogin, B., Chandu, K., Dumas, J., Elazar, Y., Hofmann, V., Jha, A.~H., Kumar, S., Lucy, L., Lyu, X., Lambert, N., Magnusson, I., Morrison, J., Muennighoff, N., Naik, A., Nam, C., Peters, M.~E., Ravichander, A., Richardson, K., Shen, Z., Strubell, E., Subramani, N., Tafjord, O., Walsh, P., Zettlemoyer, L., Smith, N.~A., Hajishirzi, H., Beltagy, I., Groeneveld, D., Dodge, J., and Lo, K.
\newblock {Dolma: An Open Corpus of Three Trillion Tokens for Language Model Pretraining Research}.
\newblock \emph{arXiv preprint}, 2024.
\newblock URL \url{https://arxiv.org/abs/2402.00159}.

\bibitem[Tan et~al.(2024)Tan, Li, Wang, Beigi, Jiang, Bhattacharjee, Karami, Li, Cheng, and Liu]{tan-etal-2024-large}
Tan, Z., Li, D., Wang, S., Beigi, A., Jiang, B., Bhattacharjee, A., Karami, M., Li, J., Cheng, L., and Liu, H.
\newblock Large language models for data annotation and synthesis: A survey.
\newblock In Al-Onaizan, Y., Bansal, M., and Chen, Y.-N. (eds.), \emph{Proceedings of the 2024 Conference on Empirical Methods in Natural Language Processing}, pp.\  930--957, Miami, Florida, USA, November 2024. Association for Computational Linguistics.
\newblock \doi{10.18653/v1/2024.emnlp-main.54}.
\newblock URL \url{https://aclanthology.org/2024.emnlp-main.54/}.

\bibitem[Trinh et~al.(2024)Trinh, Wu, Le, He, and Luong]{AlphaGeometryTrinh2024}
Trinh, T., Wu, Y., Le, Q., He, H., and Luong, T.
\newblock Solving olympiad geometry without human demonstrations.
\newblock \emph{Nature}, 2024.
\newblock \doi{10.1038/s41586-023-06747-5}.

\bibitem[Ulmer et~al.(2024)Ulmer, Mansimov, Lin, Sun, Gao, and Zhang]{ulmer2024bootstrapping}
Ulmer, D., Mansimov, E., Lin, K., Sun, J., Gao, X., and Zhang, Y.
\newblock Bootstrapping llm-based task-oriented dialogue agents via self-talk.
\newblock \emph{arXiv preprint arXiv:2401.05033}, 2024.

\bibitem[Wang et~al.(2022)Wang, Kordi, Mishra, Liu, Smith, Khashabi, and Hajishirzi]{wang2022self}
Wang, Y., Kordi, Y., Mishra, S., Liu, A., Smith, N.~A., Khashabi, D., and Hajishirzi, H.
\newblock Self-instruct: Aligning language models with self-generated instructions.
\newblock \emph{arXiv preprint arXiv:2212.10560}, 2022.

\bibitem[Wei et~al.(2022)Wei, Wang, Schuurmans, Bosma, hsin Chi, Xia, Le, and Zhou]{Wei2022ChainOT}
Wei, J., Wang, X., Schuurmans, D., Bosma, M., hsin Chi, E.~H., Xia, F., Le, Q., and Zhou, D.
\newblock Chain of thought prompting elicits reasoning in large language models.
\newblock \emph{ArXiv}, abs/2201.11903, 2022.
\newblock URL \url{https://api.semanticscholar.org/CorpusID:246411621}.

\bibitem[Xia et~al.(2024)Xia, Malladi, Gururangan, Arora, and Chen]{Xia2024LESSSI}
Xia, M., Malladi, S., Gururangan, S., Arora, S., and Chen, D.
\newblock Less: Selecting influential data for targeted instruction tuning.
\newblock \emph{ArXiv}, abs/2402.04333, 2024.
\newblock URL \url{https://api.semanticscholar.org/CorpusID:267522839}.

\bibitem[Xie et~al.(2023)Xie, Santurkar, Ma, and Liang]{xie2023data}
Xie, S.~M., Santurkar, S., Ma, T., and Liang, P.~S.
\newblock Data selection for language models via importance resampling.
\newblock \emph{Advances in Neural Information Processing Systems}, 36:\penalty0 34201--34227, 2023.

\bibitem[Yang et~al.(2023)Yang, Jia, Li, and Yan]{yang2023llm4drive}
Yang, Z., Jia, X., Li, H., and Yan, J.
\newblock Llm4drive: A survey of large language models for autonomous driving.
\newblock \emph{arXiv e-prints}, pp.\  arXiv--2311, 2023.

\bibitem[Zhang et~al.(2024)Zhang, Zeng, Hua, Ding, Chen, Ma, Li, Cui, Qi, Zhu, et~al.]{zhang2024ultramedical}
Zhang, K., Zeng, S., Hua, E., Ding, N., Chen, Z.-R., Ma, Z., Li, H., Cui, G., Qi, B., Zhu, X., et~al.
\newblock Ultramedical: Building specialized generalists in biomedicine.
\newblock \emph{arXiv preprint arXiv:2406.03949}, 2024.

\bibitem[Zhang et~al.(2023)Zhang, Zhang, Yang, Chen, Zheng, Yang, Li, Zhou, Niu, and Liu]{zhang2023gobigger}
Zhang, M., Zhang, S., Yang, Z., Chen, L., Zheng, J., Yang, C., Li, C., Zhou, H., Niu, Y., and Liu, Y.
\newblock Gobigger: A scalable platform for cooperative-competitive multi-agent interactive simulation.
\newblock In \emph{The Eleventh International Conference on Learning Representations}, 2023.

\bibitem[Zheng et~al.(2024)Zheng, Zhang, Zhang, Ye, Luo, Feng, and Ma]{zheng2024llamafactory}
Zheng, Y., Zhang, R., Zhang, J., Ye, Y., Luo, Z., Feng, Z., and Ma, Y.
\newblock Llamafactory: Unified efficient fine-tuning of 100+ language models.
\newblock In \emph{Proceedings of the 62nd Annual Meeting of the Association for Computational Linguistics (Volume 3: System Demonstrations)}, Bangkok, Thailand, 2024. Association for Computational Linguistics.
\newblock URL \url{http://arxiv.org/abs/2403.13372}.

\bibitem[Zhu et~al.(2022)Zhu, Guan, Huang, and Liu]{zhu2022storytrans}
Zhu, X., Guan, J., Huang, M., and Liu, J.
\newblock Storytrans: Non-parallel story author-style transfer with discourse representations and content enhancing.
\newblock \emph{arXiv preprint arXiv:2208.13423}, 2022.

\bibitem[Zhu et~al.(2024{\natexlab{a}})Zhu, Fu, Zhou, and Lin]{zhu2024critical}
Zhu, X., Fu, Y., Zhou, B., and Lin, Z.
\newblock Critical data size of language models from a grokking perspective.
\newblock \emph{arXiv preprint arXiv:2401.10463}, 2024{\natexlab{a}}.

\bibitem[Zhu et~al.(2024{\natexlab{b}})Zhu, Qi, Zhang, Long, Lin, and Zhou]{zhu-etal-2024-pad}
Zhu, X., Qi, B., Zhang, K., Long, X., Lin, Z., and Zhou, B.
\newblock {P}a{D}: Program-aided distillation can teach small models reasoning better than chain-of-thought fine-tuning.
\newblock In Duh, K., Gomez, H., and Bethard, S. (eds.), \emph{Proceedings of the 2024 Conference of the North American Chapter of the Association for Computational Linguistics: Human Language Technologies (Volume 1: Long Papers)}, pp.\  2571--2597, Mexico City, Mexico, June 2024{\natexlab{b}}. Association for Computational Linguistics.
\newblock \doi{10.18653/v1/2024.naacl-long.142}.
\newblock URL \url{https://aclanthology.org/2024.naacl-long.142}.

\end{thebibliography}
\bibliographystyle{icml2025}

\newpage
\appendix
\onecolumn

\section{Proof}\label{sec:proof}
\subsection{Proof of Theorem~\ref{theorem:ridgeless_fitted_linear_parameters}}

For $n=1$, we have:
\begin{align*}
\hat{w}_1 &= \tilde{X}_1^{\dagger} \tilde{Y}_1 = (X^\top X)^{-1} X^\top (Xw^* + E_1) = w^* + (X^\top X)^{-1} X^\top E_1
\end{align*}

For $n \geq 1$, we have:

\begin{align*}
\hat{w}_{n+1} &= \tilde{X}_{n+1}^{\dagger} \tilde{Y}_{n+1} \\
&= (\tilde{X}_{n+1}^\top \tilde{X}_{n+1})^{-1} \tilde{X}_{n+1}^\top \tilde{Y}_{n+1} \\
&= (X^\top X)^{-1}X^\top \tilde{Y}_{n+1}
\end{align*}

Recalling that:
\begin{align*}
\tilde{Y}_i = 
\begin{cases} 
Xw^* + E_1, & \text{if } i = 1 \\
M_{i-1}(X\hat{w}_{i-1} + E_i) + (1 - M_{i-1}) \tilde{Y}_{i-1}, 
& \text{if } 2 \leq i \leq n+1
\end{cases}
\end{align*}

Substituting this $\tilde{Y}_i$ into the expression for $\hat{w}_{n+1}$:

We begin the data editing data process:
\begin{align}
\tilde{Y}_2 = M_1(X \hat{w}_1 + E_2) + (1 - M_1) \tilde{Y}_1
\end{align}

Then:
\begin{align}
\tilde{Y}_3 = M_2(X \hat{w}_2 + E_3) + (1 - M_2) \tilde{Y}_2
\end{align}

We have: 
\begin{align*}
\tilde{Y}_3 &= M_2(X \hat{w}_2 + E_3) + (1 - M_2) \left( M_1(X \hat{w}_1 + E_2) + (1 - M_1) \tilde{Y}_1 \right) \\
& = M_2(X \hat{w}_2 + E_3) + (1 - M_2)M_1(X \hat{w}_1 + E_2) + (1-M_2)(1 - M_1) \tilde{Y}_1 
\end{align*}

We can expand \(\tilde{Y}_{n+1}\) by recursively substituting the previous expressions:

\begin{align}
\tilde{Y}_{n+1} &= M_n(X \hat{w}_n + E_{n+1}) + (1 - M_n) \tilde{Y}_n \\
&= M_n(X \hat{w}_n + E_{n+1}) + (1 - M_n) \left[ M_{n-1}(X \hat{w}_{n-1} + E_n) + (1 - M_{n-1}) \tilde{Y}_{n-1} \right] \\
&= M_n(X \hat{w}_n + E_{n+1}) + (1 - M_n)M_{n-1}(X \hat{w}_{n-1} + E_n) + (1 - M_n)(1 - M_{n-1}) \tilde{Y}_{n-1} \\
&\vdots \\
&= \sum_{i=1}^{n} \left[ \left( \prod_{j=i+1}^{n} (1 - M_j) \right) M_i (X \hat{w}_i + E_{i+1}) \right] + \left( \prod_{j=1}^{n} (1 - M_j) \right) \tilde{Y}_1
\end{align}

Recalling properties of $M_i$:
\begin{align}
    M_i(1 - M_i) &= 0 \quad \text{and} \quad (1 - M_i) M_i = 0 \\
    M_i M_j &= 0 \quad \text{for} \quad i \neq j \\
    (1 - M_i)(1 - M_j) &= 1 - M_i - M_j \quad \text{for} \quad i \neq j \\
\end{align}

Then we have:
\begin{align}
    \tilde{Y}_{n+1} &= \sum_{i=1}^{n} M_i (X \hat{w}_i + E_{i+1}) + \left( 1 - \sum_{i=1}^{n} M_i \right) \tilde{Y}_1 \\
    &= \sum_{i=1}^{n} M_i (X \hat{w}_i + E_{i+1}) + \left( 1 - \sum_{i=1}^{n} M_i \right) (Xw^*+E_1) \\
    &= X w^* + E_1 + \sum_{i=1}^{n} M_i \left( X (\hat{w}_i - w^*) + (E_{i+1} - E_1) \right)
\end{align}

Substituting this back into the expression for $\hat{w}_{n+1}$:
\begin{align}
\hat{w}_{n+1} 
& = (X^\top X)^{-1} X^\top \left[ X w^* + E_1 + \sum_{i=1}^{n} M_i \left( X (\hat{w}_i - w^*) + (E_{i+1} - E_1) \right) \right] \\
& = w^* + (X^\top X)^{-1} X^\top \left[  E_1 + \sum_{i=1}^{n} M_i X (\hat{w}_i - w^*) + \sum_{i=1}^{n} M_i (E_{i+1} - E_1) \right]
\end{align}

We can observe:
\begin{align}
\hat{w}_1 &= (X^\top X)^{-1} X^\top (X w^* + E_1) = w^* + (X^\top X)^{-1} X^\top E_1 \\
\hat{w}_2 &= w^* + (X^\top X)^{-1}  X^\top \left( M_1 X (X^\top X)^{-1} X^\top E_1 + M_1 E_2 + (1 - M_1) E_1 \right) \\
&= w^* + (X^\top X)^{-1}  X^\top \left( E_1 + M_1 E_2 \right)
\end{align}

We prove this Theorem~\ref{theorem:ridgeless_fitted_linear_parameters} by induction. 

\paragraph{Inductive Step:} Assume the formula holds for $n$, we have:
\begin{align}
\hat{w}_{n+1} &= w^* + (X^\top X)^{-1} X^\top \left( E_1 + M_1 E_2 + M_2 E_3 + \dots + M_{n} E_{n+1} \right) \\
&= w^* + (X^\top X)^{-1} X^\top \left( E_1 + \sum_{i=1}^{n} M_i E_{i+1} \right)\label{eq:weight_update}
\end{align}

Substitute \( \hat{w}_{i} \) into \( \hat{w}_{n+1} \):

Then we can get:
\begin{align}
\hat{w}_{n+1} 
&= w^* + (X^\top X)^{-1} X^\top \left[ E_1 + \sum_{i=1}^{n} M_i P \left( E_1 + \sum_{j=1}^{i-1} M_j E_{j+1} \right) + \sum_{i=1}^{n} M_i (E_{i+1} - E_1) \right] \\
&= w^* + (X^\top X)^{-1} X^\top \left[ E_1 + \sum_{i=1}^{n} M_i  \left( E_{i+1} + \sum_{j=1}^{i-1} M_j E_{j+1} \right) \right] \\
&= w^* + (X^\top X)^{-1} X^\top \left( E_1 + \sum_{i=1}^{n} M_i E_{i+1} \right) \\
&\text{where}\quad P = X(X^\top X)^{-1} X^\top,
\end{align}
The above derivation aligns with Theorem~\ref{theorem:ridgeless_fitted_linear_parameters}, and the proof is complete.

\subsection{Proof of Theorem~\ref{theorem:upper_bound}}
We substitute the Eq.~\ref{eq:weight_update} into Test Error Eq.~\ref{eq:test_error}:
\begin{align}
E_{test}(\hat{w}_{n+1}) & = \mathbb{E}\left[\left\| (X^\top X)^{-1} X^\top \left( E_1 + \sum_{i=1}^{n} M_i E_{i+1} \right) \right\|_{\Sigma}^2 \right] \\
& = \mathbb{E}\left[ \left( E_1 + \sum_{i=1}^{n} M_i E_{i+1} \right)^\top X (X^\top X)^{-2} X^\top \left( E_1 + \sum_{i=1}^{n} M_i E_{i+1} \right) \right] \\
& = \sigma^2 \mathbb{E}\left[\text{tr}\left((X^\top X)^{-1}\right)\right] + \sigma^2 \sum_{i=1}^{n} \mathbb{E}\left[\text{tr}\left(M_i (X^\top X)^{-1} M_i\right)\right] \\
& = \sigma^2 \mathbb{E}\left[\text{tr}\left((X^\top X)^{-1}\right)\right] + \sigma^2 \sum_{i=1}^{n} \mathbb{E}\left[\text{tr}\left((X^\top X)^{-1} M_i\right)\right]\label{eq:relaxed_term}
\end{align}

Further, by applying the Cauchy-Schwarz inequality~\citep{rudin1976principles}, we obtain:
\begin{align}
E_{test}(\hat{w}_{n+1}) 
& \leq \sigma^2 \mathbb{E}\left[\text{tr}\left((X^\top X)^{-1}\right)\right] + \sigma^2 \sqrt{\mathbb{E}\left[\text{tr}\left((X^\top X)^{-2}\right)\right]} \cdot \sum_{i=1}^{n}\sqrt{\mathbb{E}\left[\text{tr}(M_i)\right]} 
\end{align}

We refer to the following lemma~\citep{dohmatob2024model}, which is essential for proving Theorem 2:
\begin{lemma}
Let $T$ and $d$ be positive integers with $T \geq d + 2$, and let $X \in \mathbb{R}^{T \times d}$ be a random matrix with i.i.d. rows from $\mathcal{N}(0, \Sigma)$ with $\Sigma$ positive definite. Then, $X$ has full rank a.s. Moreover, it holds that:
\begin{align}
\mathbb{E}_X\left[(X^\top X)^{-1}\right] = \frac{1}{T - d - 1} \Sigma^{-1}.
\end{align}
\label{lemma:tr_inv_cov_eq_prefactor_inv_cov}
\end{lemma}

Using Lemma~\ref{lemma:tr_inv_cov_eq_prefactor_inv_cov}, we have:

\begin{align}
E_{test}\left[\text{tr}\left( (X^\top X)^{-1} \right)) \right] & = \frac{d}{T - d - 1}
\end{align}

Then, we have:
\begin{align}
E_{test}(\hat{w}_{n+1}) &= \sigma^2 \mathbb{E}\left[\text{tr}\left((X^\top X)^{-1}\right)\right] + \sigma^2 \sum_{i=1}^{n} \mathbb{E}\left[\text{tr}\left((X^\top X)^{-1} M_i\right)\right] \\
 & \leq \frac{\sigma^2d}{T - d - 1} + \sigma^2 \sqrt{\mathbb{E}\left[\text{tr}\left((X^\top X)^{-2}\right)\right]} \cdot \sum_{i=1}^{n}\sqrt{\mathbb{E}\left[\text{tr}(M_i)\right]} 
\end{align}

In our setting, the data is incrementally modified over iterations and modifications decreases progressively. This behavior can be modeled by the sum of a geometric series, where the amount of modified data decreases by a fixed ratio \(\eta\) with each iteration. Then, we assume the editing operation as 
$||M_i|| = ||M_{i-1}||\eta, \text{for}\ i=1,2,\dots,n.$
Therefore, the test error for data editing can be bounded:

\begin{align}
E_{test}(\hat{w}_{n+1}) \leq \frac{\sigma^2d}{T - d - 1} + \sigma^2 \sqrt{\mathbb{E}\left[\text{tr}\left((X^\top X)^{-2}\right)\right]} \cdot \frac{\sqrt{\mathbb{E}\left[\text{tr}(M_1)\right]}}{1 - \eta}
\end{align}

Additionally, since \(M_i\) is not full-rank, as seen from Eq.~\ref{eq:relaxed_term}, we can apply a more relaxed and simplified bound, as follows:
\begin{align}
E_{test}(\hat{w}_{n+1}) \leq \frac{2\sigma^2d}{T - d - 1} 
\end{align}

Thus, the above derivation satisfies the Theorem~\ref{theorem:upper_bound}.

\begin{algorithm}
\caption{Token-level Editing}
\begin{algorithmic}[1]
\STATE \textbf{Input:} Sequence of tokens $\textbf{x} = (x_1, \dots, x_t)$, prior distribution $P$, probability threshold $p$
\STATE \textbf{Output:} Edited sequence $\textbf{x'} = (x_1', \dots, x_t')$

\FOR{each token $x_i$ in sequence $\textbf{x}$}
    \STATE Compute conditional probability $P(x_i \mid x_1, \dots, x_{i-1})$
    \IF{$P(x_i \mid x_1, \dots, x_{i-1}) \geq p$}
        \STATE Resample token $\tilde{x}_i$ from prior distribution
        \STATE Set $x_i' \gets \tilde{x}_i$
    \ELSE
        \STATE Set $x_i' \gets x_i$
    \ENDIF
\ENDFOR

\STATE \textbf{Return:} Edited sequence $\textbf{x'} = (x_1', \dots, x_t')$
\end{algorithmic}
\end{algorithm}

\begin{table}[htbp]
\centering
\caption{
Comparison of human and synthetic data performance across downstream tasks in~\citep{Maini2024RephrasingTW},  based on training with GPT-2.
}\label{tab:human_vs_synthetic_downstream_tasks}
\resizebox{0.9\textwidth}{!}{
\begin{tabular}{l|ccccccc|c}
\toprule
                    & TruthfulQA & LogiQA & Wino. & PIQA  & ARC-E & BoolQ & OBQA & Avg             \\ 
\midrule
Human Data         & 32.68      & 23.03  & 51.3  & 64.42 & 44.4  & 60.98 & 15   & 41.69  \\ 
25\% Synthetic Data & 27.91      & 21.37  & 50.12 & 63.93 & 43.94 & 62.29 & 15.4 & 40.71  \\ 
50\% Synthetic Data & 30.84      & 22.58  & 52.41 & 63.33 & 44.02 & 62.14 & 16   & 41.62 \\ 
75\% Synthetic Data & 29.5       & 22.65  & 49.8  & 63.44 & 44.53 & 61.56 & 17.2 & 41.24  \\ 
Synthetic Data     & 28.89      & 22.58  & 49.72 & 63    & 46.3  & 54.53 & 16.8 & 40.26  \\ 
\bottomrule
\end{tabular}
    } 
\end{table}

\begin{figure}[t]
    \centering
    \includegraphics[width=0.5\linewidth]{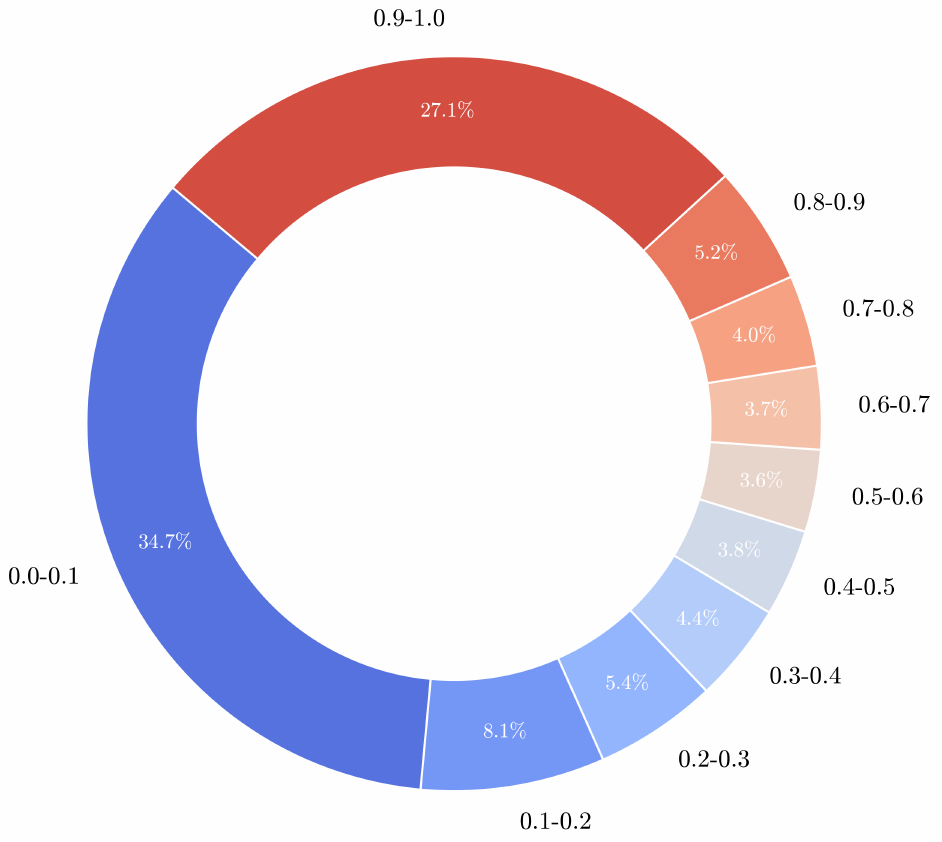}
    \caption{Token distribution across different probability ranges in BioMed dataset.}
    \label{fig:token_distribution}
\end{figure}

\begin{figure}[t]
    \centering
    \includegraphics[width=0.7\linewidth]{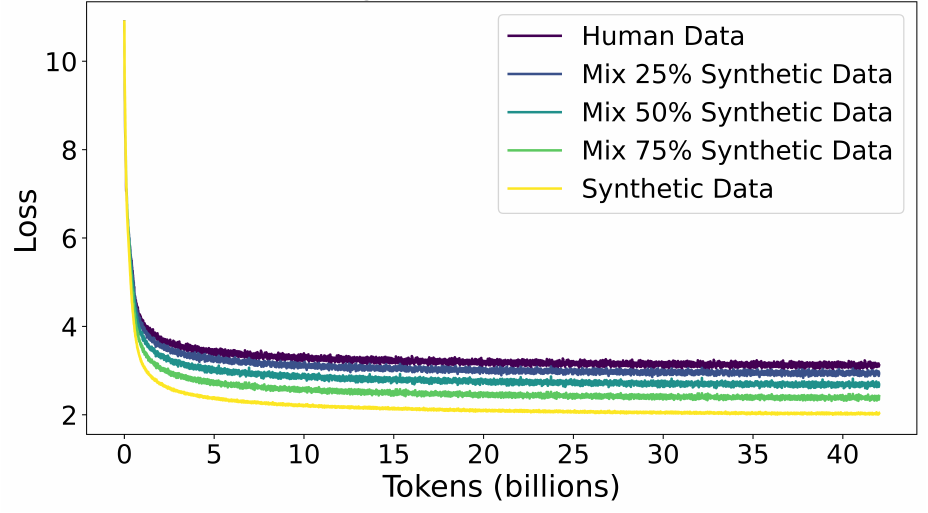}
    \caption{Pre-training loss of GPT-2 Small (124M) on human (Dolma~\citep{dolma}) and synthetic (Cosmopedia~\citep{benallal2024cosmopedia}) data. As the proportion of synthetic data increases, the model’s loss decreases.}\label{fig:training_loss_synthetic_data}
\end{figure}

\section{More Related Work}\label{sec:more_related_work}
Phi-1/2~\cite{gunasekar2023textbooks} demonstrate that the synthetic data can boost training efficiency and performance compared to raw data in language model pre-training. Furthermore, \citet{feng2024beyond} introduce a verifier to filter synthetic samples, theoretically avoiding model collapse. \citet{liu2024best,tan-etal-2024-large} highlight that synthetic data will play a crucial role in the development of AI. For example, synthetic data can be used to construct highly specialized datasets, enhancing the performance of downstream tasks. \citet{AlphaGeometryTrinh2024} utilize synthetic math data to train a 125M language model, which successfully solved 25 out of 30 selected problems from the International Mathematical Olympiad (IMO) problem set. \citet{zhang2024ultramedical} develop a biomedical instruction dataset that was used to train specialized bio-models, enabling them to excel in answering questions related to medical exams and clinical scenarios. \citet{eldan2023tinystories} introduce a novel synthetic dataset and evaluation paradigm that enables small language models to generate coherent, diverse, and grammatically sound stories. As outlined above, in the post-training stages of LLMs, synthetic data enhances downstream task performance and aligns foundation models with humans. And~\citet{Maini2024RephrasingTW} propose rephrasing the pre-training data into a Wikipedia or Q/A style to achieve better alignment with downstream tasks. Synthetic data is a powerful tool for training. Our approach is also based on synthetic data methods. Instead of sampling data solely based on this prior, we modify the data using the prior as a guide.

\citet{bertrand2023stability} develop a rigorous framework to demonstrate the importance of real data in maintaining the stability of iterative training. \citet{ferbach2024self} theoretically demonstrate that the impact of data curation can be formalized as an implicit preference optimization mechanism. \citet{kazdan2025collapse} reveal the detailed training dynamics of model collapse under three different training workflows. Of course, there are also some remarkable studies that successfully used synthetic data. \citet{wang2022self} propose the Self-Instruct data generation framework, enhancing instruction-following capabilities. \citet{ulmer2024bootstrapping} employ the self-talk method to generate high-quality data. ReST~\cite{gulcehre2023reinforced} uses a policy model to generate datasets and then employs offline RL to fine-tune LLMs on generated datasets. \citet{singh2023beyond} demonstrate that self-training with binary feedback filtering can reduce reliance on real data. \citet{alemohammad2023self} demonstrate that without enough fresh real images, future generative models will gradually decline.  \citet{briesch2023large} illustrates that real data in the iterative training process can slow the decline of LLMs, but cannot fully prevent it. \citet{martinez2023towards} shows that the quality and diversity of generated images degrade over time.

\section{More Discussion of Main Results}\label{sec:More_Discussion_of_Main_Results}
As shown in Table~\ref{tab:cpt}, our method shows consistent improvements over the source data across OLMo-1B and LLaMA-3-8B. For instance, in the Biomedicine domain, the average score for OLMo-1B increased from 36.63 to 40.89 with ToEdit, while LLaMA-3-8B saw an increase from 54.13 to 56.48. Table~\ref{tab:pt} further supports the effectiveness of our approach in pre-training. The average performance of OLMo-1B increases from 32.75 to 33.11, reflecting improved generalization capabilities. While the improvement is modest, the consistent trend across tasks like PIQA, BoolQ, and ARC-c highlights the broader applicability of our method. As for SFT results in Table~\ref{tab:sft}, using both the original and edited data, the results indicate a small but consistent improvement. Specifically, ToEdit improves original FLAN v2, with average performance increasing from 70.18 to 70.65. For Natural Instructions, the average performance of LLaMA-3-8B improves from 69.34 to 69.70, with gains in tasks like Winogrande and SIQA. These improvements demonstrate the adaptability of our method to instruction-tuning tasks. For code-related tasks, the improvements demonstrate better reasoning and code comprehension.

\section{Comparison with Pure Synthetic Data and Reformat Methods}\label{sec:Explanation_of_Synthetic_Data}
\paragraph{Definition and Characteristics of Synthetic Data}
Synthetic data ($D_s$) can be categorized based on its relationship with the distributions of a language model ($P_{\text{LM}}$) and human-produced data ($P_{\text{data}}$) during the generation process, quantified as $d = \text{KL}( \cdot || P_{\text{data}})$:
\begin{align}
D_s = 
\begin{cases} 
D_s^{\text{pure}} \sim P_{\text{LM}}, & \text{if }  \text{KL}(P_{\text{LM}} || P_{\text{data}}) > \epsilon, \\
D_s^{\text{semi}} \sim P_{\text{semi}}, & \text{if }  \text{KL}(P_{\text{semi}} || P_{\text{data}}) \leq \epsilon.
\end{cases}
\end{align}
where \textbf{Pure Synthetic Data $D_s^{\text{pure}}$:} Generated entirely from the language model ($D_s^{\text{pure}} \sim P_{\text{LM}}$), with a KL divergence $\text{KL}(P_{\text{LM}} \| P_{\text{data}})$ exceeding a threshold $\epsilon$. This implies a significant deviation of the language model’s distribution from the human-produced data distribution. \textbf{Semi-Synthetic Data $D_s^{\text{semi}}$:} Derived from limited modifications to human-produced data ($P_{\text{data}}$), ensuring that the resulting distribution ($P_{\text{semi}}$) has a KL divergence $\text{KL}(P_{\text{semi}} || P_{\text{data}})$ bounded by $\epsilon$. This reflects a closer alignment of semi-synthetic data with human-produced data.

From the generation process, pure synthetic data $D_s^{\text{pure}}$: This data is induced by a language model through prompts and does not modify human-produced data, resulting in low overlap content with human-produced data. For example, Cosmopedia~\citep{benallal2024cosmopedia} expands human-produced data and generates data without human-produced data. Semi-Synthetic Data $D_s^{\text{semi}}$: This data is generated by directly modifying human-produced data, such as paraphrasing or token-level editing. It derives from transformations of human-produced data. For example, WRAP~\citep{Maini2024RephrasingTW} generates paraphrases of human-produced data. ToEdit (ours) performs token editing on human-produced data.

Specifically, both \textit{Rephrasing the Web}~\citep{Maini2024RephrasingTW} and our token-level editing aim to refine data while preserving the original distribution, producing semi-synthetic data.
In contrast, purely synthetic data in Cosmopedia lacks the long-tail distribution and overly concentrates on n-gram features. Ultimately, semi-synthetic data enhances training performance, whereas purely synthetic data results in model collapse. Moreover, replacing a whole real sample with synthetic data can damage the performance.

The primary distinction between Cosmopedia, Rephrasing the Web~\citep{Maini2024RephrasingTW}, and our approach lies in how much of the original human data distribution is preserved. We provide a detailed comparison of these synthetic methods in Table~\ref{tab:methods}. 
\begin{table}[ht]
  \centering
  \caption{Comparison of different synthetic data methods.}
  \resizebox{\textwidth}{!}{
    \begin{tabular}{l|l|p{7cm}|l}
\toprule
Method & Data Type & Approach & Result \\
\midrule
Cosmopedia~\citep{benallal2024cosmopedia} & Pure synthetic & Using a prompt to induce data from LLMs. & Reveal non-iterative model collapse. \\ 
Rephrasing the Web~\citep{Maini2024RephrasingTW} & Semi-synthetic & Using a prompt and source content to guide LLMs to reformat source content. & Improve training performance. \\ 
ToEdit (Ours) & Semi-synthetic & Using the distribution of source content estimated by LLMs (single forward pass) to replace tokens. & Improve training performance. \\ 
\bottomrule
\end{tabular}
}
\label{tab:methods}
\end{table}

\section{More Results of Human and Synthetic Data Mixture Training}\label{sec:more_pretraining_results}
We provide more training results for the human and synthetic data mixture.
The main results and analysis can be found in Sec~\ref{sec:non_iterative_model_collapse}. Except for GPT-2 pre-training, we also use the OLMo models~\citep{OLMo} for further experiments. 

\begin{figure*}[t]
    \centering
    \begin{minipage}[t]{0.48\textwidth}
        \centering
        \includegraphics[width=\linewidth]{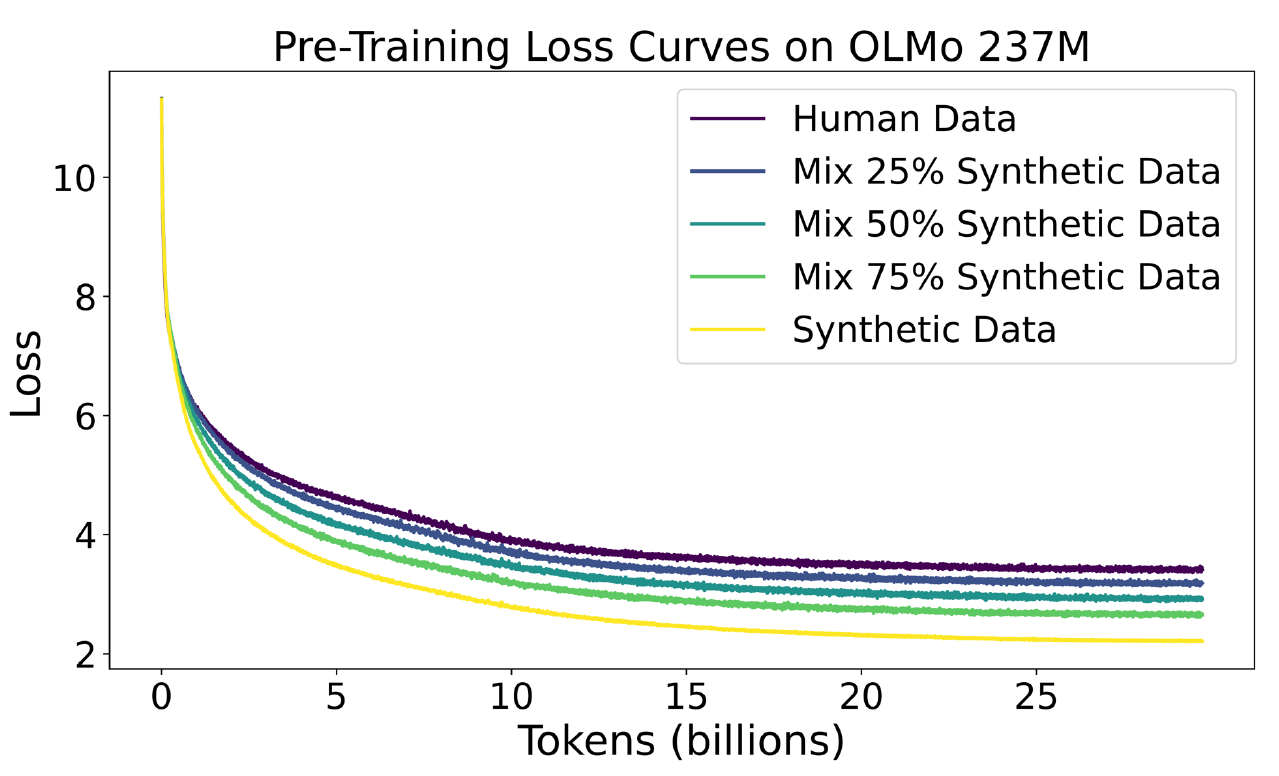}
        \caption{OLMo-237M pretraining with mixed human and synthetic data proportions. We pretrain the OLMo-237M model using a mixture of human data (Dolma~\citep{dolma}) and synthetic data (Cosmopedia~\citep{benallal2024cosmopedia}).}
        \label{fig:olmo_pretrainig}
    \end{minipage}%
    \hfill 
    \begin{minipage}[t]{0.48\textwidth}
        \centering
        \includegraphics[width=\linewidth]{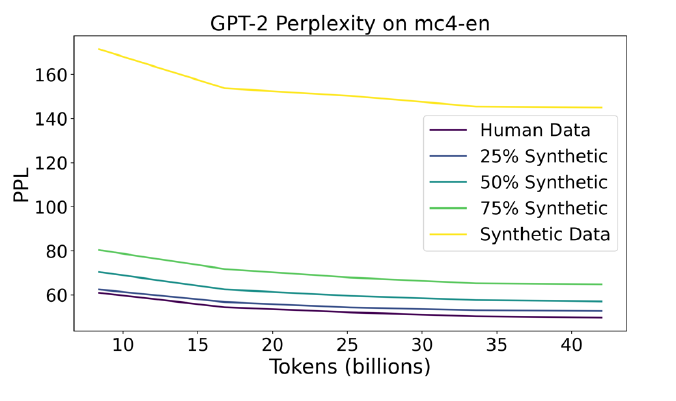}
        \caption{GPT-2 perplexity (PPL) on validation sets, trained from scratch.}
        \label{fig:mc4_ppl_gpt2}
    \end{minipage}
\end{figure*}

As shown in Figure~\ref{fig:olmo_pretrainig}, the training loss continues to decrease as the amount of synthetic data increases, which is consistent with GPT-2 pre-training in Figure~\ref{fig:gpt2-pretraining-and-eval}. More synthetic data can lead to better fitting. However, a lower loss does not necessarily mean a better model. As illustrated in Figure~\ref{fig:gpt2-pretraining-and-eval}B and~\ref{fig:mc4_ppl_gpt2}, models that fits better perform worse in real world tasks.

Furthermore we follow~\cite{Maini2024RephrasingTW} to conduct more experiments including PPL results on 22 validation sets of Pile~\citep{Gao2020ThePA} and general understanding tasks. The additional results in Table~\ref{tab:human_vs_synthetic_downstream_tasks},~\ref{tab:human_vs_synthetic_downstream_tasks_olmo} and~\ref{tab:ppl_results_of_pile} are consistent with our findings. Specifically, the PPL increases as the proportion of purely synthetic data grows, while the performance on downstream tasks similarly exhibits a gradual decline with the increase in synthetic data.

\begin{table}[htbp]
\centering
\caption{
Comparison of human and synthetic data performance across downstream tasks in~\citep{Maini2024RephrasingTW},  based on training with OLMo-237M. ± indicates the standard error.
}\label{tab:human_vs_synthetic_downstream_tasks_olmo}
\resizebox{0.9\textwidth}{!}{
\begin{tabular}{l|cccccc|c}
\toprule
                    & TruthfulQA & LogiQA & Wino. & PIQA  & ARC-E &  OBQA & Avg  \\ 
\midrule
Human Data            & 26.81 $\pm$ 1.550  & 21.06 $\pm$ 1.028  & 52.01 $\pm$ 1.404  & 56.69 $\pm$ 1.156  & 31.73 $\pm$ 0.9550  & 13.80 $\pm$ 1.543  & 33.68 \\
25\% Synthetic Data   & 26.44 $\pm$ 1.543  & 21.25 $\pm$ 1.032  & 52.64 $\pm$ 1.403  & 57.02 $\pm$ 1.155  & 31.78 $\pm$ 0.9552  & 12.40 $\pm$ 1.475  & 33.59 \\
50\% Synthetic Data   & 25.95 $\pm$ 1.534  & 20.04 $\pm$ 1.099  & 52.25 $\pm$ 1.408  & 56.64 $\pm$ 1.126  & 31.82 $\pm$ 0.9557  & 12.80 $\pm$ 1.495  & 33.25 \\
75\% Synthetic Data   & 25.34 $\pm$ 1.522  & 20.87 $\pm$ 1.025  & 50.43 $\pm$ 1.405  & 55.60 $\pm$ 1.159  & 32.74 $\pm$ 0.9629  & 12.00 $\pm$ 1.454  & 32.83 \\
Synthetic Data        & 23.01 $\pm$ 1.473  & 20.29 $\pm$ 1.014  & 49.33 $\pm$ 1.405  & 55.93 $\pm$ 1.158  & 33.33 $\pm$ 0.9673  & 14.20 $\pm$ 1.562  & 32.68 \\
\bottomrule
\end{tabular}
    } 
\end{table}

\section{Experiment Settings}\label{sec:experiment_setting}
In this section, we describe our experiments settings in detail.
\subsection{Training}
\paragraph{Pre-training} We utilized both GPT-2 and OLMo models. The pre-training datasets included Dolma, representing real data, and Cosmopedia, representing synthetic data. For GPT-2, we employed the official FSDP (Fully Sharded Data Parallel) framework provided by Torch for training. For OLMo\footnote{\url{https://github.com/allenai/OLMo}}, we used the official open-source computational code, which also incorporates the FSDP framework alongside Flash Attention for acceleration.



\begin{table}[htbp]
\centering
\caption{Case Study.}
\label{tab:case_study}
\begin{tabular}{p{0.3\linewidth}p{0.3\linewidth}p{0.2\linewidth}}
\toprule
\textbf{Before (source)} & \textbf{After (edited)} & \textbf{Changes} \\
\midrule
Construct a function using PHP language that applies lexical analysis on a provided text string to analyze the individual, non-repeated words elements present. & Construct a function using PHP language that applies lexical analysis on a provided text string to quantify unique words. & ``analyze'' $\rightarrow$ ``quantify'' \\
\midrule
Test with provided string, \texttt{\$str = 'Greetings, Planet Earth!'}. & Test with provided string, \texttt{\$str = 'Greetings, Planet Earth!'}. & No changes. \\
\midrule
Implements \texttt{wordCount} to remove punctuation, convert text to lowercase, split into words, and count unique words. & Implements \texttt{wordCount} to remove punctuation, convert text to lowercase, split into words, and calculate unique words. & ``count'' $\rightarrow$ ``calculate'' \\
\midrule
Returns \texttt{\{'greetings': 1, 'planet': 1, 'earth': 1\}}. & Returns \texttt{\{'greetings': 1, 'planet': 1, 'earth': 1\}}. & No changes. \\
\bottomrule
\end{tabular}
\end{table}

\paragraph{Continual Pre-training} We follow~\cite{cheng2024adapting} to conduct continual pre-training on biomedicine, finance, and math domains. Specifically, PubMed Abstracts from the Pile are utilized as the pre-training corpora for the biomedicine  domain. For the finance domain, financial news data covering over 7,000 stocks from May 2022 to May 2023 is collected using the FinGPT framework. We continue pre-training OLMo-1B and LLaMA-3-8B on each domain. For implementation, we utilized the official training framework for OLMo-1B, leveraging Fully Sharded Data Parallel (FSDP) for continual pre-training. For LLaMA, we adopted the LLaMA-Factory framework to carry out the continual pre-training process. Our experiments was primarily conducted on OLMo-1B and Llama-3-8B models, with Llama-3-8B utilizing LoRA (Low-Rank Adaptation) for parameter-efficient fine-tuning. The data and evaluation are given in this repo\footnote{\url{https://github.com/microsoft/LMOps/tree/main/adaptllm}}. We conducted the continual pre-training on a total of 1B tokens.

\paragraph{Supervised Fine-tuning} We used the Llama-Factory~\citep{zheng2024llamafactory} framework to fine-tune Llama-3-8B. As for general instruction tuning tasks, we adopt instruction tuning datasets from~\citep{Xia2024LESSSI}~\footnote{\url{https://huggingface.co/datasets/princeton-nlp/less_data}}, including CoT~\citep{Wei2022ChainOT} , FLAN v2~\citep{longpre2023flan}, and Open Assistant 1~\citep{Kopf2023OpenAssistantC}. As for code-related reasoning tasks, we utilize OSS-Instruct-75K~\footnote{\url{https://huggingface.co/datasets/ise-uiuc/Magicoder-OSS-Instruct-75K}} and Evol-Instruct-110K~\footnote{\url{https://huggingface.co/datasets/ise-uiuc/Magicoder-Evol-Instruct-110K}}. These datasets provide sufficient diversity for verification on fine-tuning. We apply LoRA~\cite{hu2021lora} to Llama-3-8B experiments.


\subsection{Evaluation}
\paragraph{Pre-training} We use PPL and downstream tasks to conduct analysis and performance test. As for PPL, it stands for perplexity, a commonly used metric in NLP to evaluate the quality of language models. It measures how well a probabilistic model predicts a given dataset, with lower values indicating better performance. Formally, the perplexity of a language model is calculated as:
\[
\text{PPL} = 2^{-\frac{1}{N} \sum_{i=1}^{N} \log_2 P(x_i)}
\]
Alternatively, it can also be expressed as:
\[
\text{PPL} = \exp\left(-\frac{1}{N} \sum_{i=1}^{N} \log P(x_i)\right)
\]
Where $N$ is the number of tokens in the dataset, and $P(x_i)$ is the predicted probability of the $i$-th token. Perplexity essentially represents the exponential of the average negative log-likelihood of the predicted tokens, indicating how “perplexed” the model is when making predictions. 

As for downstream tasks, we use general understanding tasks in~\citep{Maini2024RephrasingTW} to analyze model collapse in Table~\ref{tab:human_vs_synthetic_downstream_tasks} and general test tasks in~\citep{Cheng2024InstructionPL} to test our methods in Table~\ref{tab:pt}. All downstream tasks we used can be found in~\citep{eval-harness}\footnote{\url{https://github.com/EleutherAI/lm-evaluation-harness}}.

\paragraph{Continual Pre-training} We use the test data and code in~\citep{cheng2024adapting}\footnote{\url{https://github.com/microsoft/LMOps/tree/main/adaptllm}} to test domain specific task performance after CPT.
\paragraph{Supervised Fine-tuning} We utilize the general downstream tasks from~\citep{Cheng2024InstructionPL} to evaluate instruction-tuning performance and reasoning tasks to assess reasoning capabilities. All downstream tasks we used can be found in~\citep{eval-harness}\footnote{\url{https://github.com/EleutherAI/lm-evaluation-harness}}.



\begin{table}[htbp]
  \centering
  \caption{PPL results of GPT-2 124M pre-training on mixture of human and synthetic data.}
  \resizebox{\textwidth}{!}{
    \begin{tabular}{l|lllll|lllll|lllll}
\toprule
Synthetic Data Ratio & \multicolumn{5}{c|}{25\%} & \multicolumn{5}{c|}{50\%} & \multicolumn{5}{c}{75\%} \\
\midrule
Tokens Size & 8.4B & 16.8B & 25.2B & 33.6B & 42B & 8.4B & 16.8B & 25.2B & 33.6B & 42B & 8.4B & 16.8B & 25.2B & 33.6B & 42B \\
\midrule
Epochs & 1 & 2 & 3 & 4 & 5 & 1 & 2 & 3 & 4 & 5 & 1 & 2 & 3 & 4 & 5 \\
\midrule
Wikitext-103 & 45.97 & 39.87 & 37.65 & 36.91 & 36.32 & 50.29 & 43.15 & 40.46 & 39.43 & 38.65 & 58.66 & 48.75 & 45.20 & 43.42 & 42.95 \\
RedPajama & 42.28 & 37.62 & 35.72 & 34.66 &  34.24 & 46.89 & 41.42 & 39.37 & 38.21 & 37.72 & 55.72 & 49.26 & 46.27 & 44.81 & 44.30 \\
Falcon-RefinedWeb & 56.40 & 50.62 & 48.26 & 47.13 & 46.66 & 61.06 & 54.34 & 51.72 & 50.39 & 49.87 &  69.32 & 61.50 & 58.28 & 56.77 & 56.19 \\
c4-en & 48.15 & 43.14 & 40.98 & 39.91 & 39.41 & 51.79 & 46.06 &  43.90 & 42.73 & 42.23 & 58.60 & 52.22 & 49.26 & 47.87 & 47.27 \\
mc4-en & 62.46 & 56.80 & 54.35 & 53.06 & 52.71 & 70.43 & 62.48 & 59.61 & 57.66 & 57.07 & 80.37 & 71.77 &  67.90 & 65.31 & 64.82 \\
\bottomrule
  \end{tabular}}
  \label{tab:data_mixture}
\end{table}

\begin{table}[htbp]
  \centering
  \caption{PPL results of OLMo-237M  pretraining on mixture of human and synthetic data.}
  \resizebox{\textwidth}{!}{
    \begin{tabular}{l|lllll|llllllllll}
\toprule
Synthetic Data Ratio & 0\% & 25\% & 50\% & 75\% & 100\% & DSIR (1M) & DSIR (10M) & Edu Classifier (1M)& Edu Classifier (10M) & PPL Filter (1M) & PPL Filter (10M) & Density Sampling (1M) & Density Sampling (10M)\\
\midrule
Unique Tokens & 8.4B & 8.4B & 8.4B & 8.4B &  8.4B & 0.6B & 8.4B & 0.75B & 7.4B & 0.97B & 9B & 0.6B & 7.1B \\
Training Tokens & 8.4B & 8.4B & 8.4B & 8.4B & 8.4B & 8.4B & 8.4B & 10.5B & 7.4B & 13.68B& 9B & 8.9B & 7.1B\\
Epochs & 1 & 1 & 1 & 1 & 1 & 14 & 1 & 14 & 1 & 14 & 1 & 14 & 1\\
\midrule
Wikitext-103 & 187.36 & 185.5 &  260.08 & 367.46 & 1605.73 & 1309.53 & 1757.03 & 1111.29 & 1612.95 & 738.36 & 1193.25 & 1188.40 & 1753.89 \\
RedPajama  & 175.38 & 183.93 & 236.33 & 301.09 & 907.91 & 649.36 & 916.51 & 811.14 & 1104.75 & 376.36 & 645.82 & 789.67 & 896.18\\
Falcon-RefinedWeb & 165.17 & 166.69 &  199.68 & 245.15 & 523.93 & 573.61 & 510.96 & 522.97 & 612.72 & 344.82 & 449.86 & 501.99 & 560.92\\ 
c4-en & 123.88 & 127.68 & 147.69 & 174.48 & 410.19 & 457.96 & 404.63 & 415.88 & 487.97 & 286.95 & 367.44 & 414.55 & 457.71\\
mc4-en & 208.91 & 208.94 & 263.35 & 324.91 & 800.40 & 861.01 & 823.12 & 769.86 & 955.70 & 476.81 & 662.00 & 740.75 & 844.53\\
M2D2-Wiki & 88.24 &  87.34 & 107.77 & 114.19 & 189.06 & 234.45 & 183.17 & 161.58 & 206.45 & 130.43 & 162.08 & 167.20 & 205.50 \\
M2D2-S2ORC & 86.15 & 81.53 & 97.61 & 100.64 & 204.22 & 170.78 & 496.40 & 145.27 & 201.52 & 117.44 & 163.38 & 131.22 & 192.97\\

\bottomrule
  \end{tabular}}
  \label{tab:olmo-237m}
\end{table}

\begin{figure*}[t]
\small
  \centering
  \includegraphics[width=\linewidth]{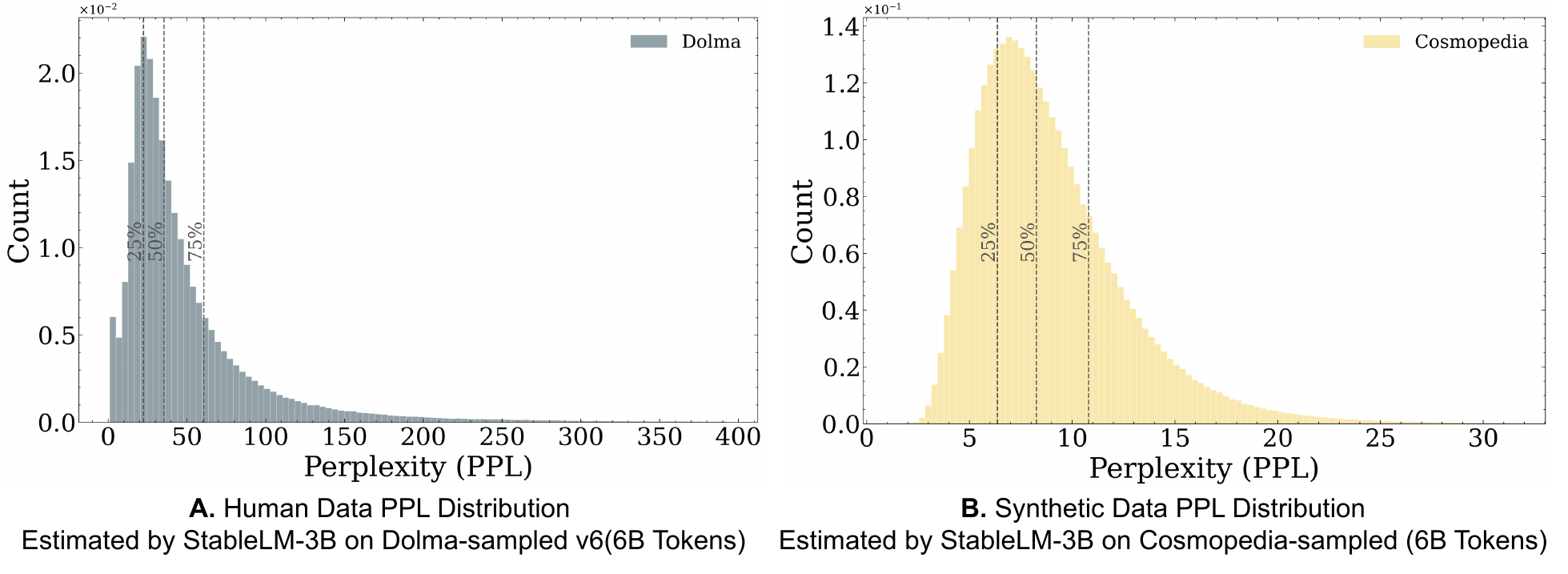}
  \vspace{-2em}
  \caption{PPL distribution of human and synthetic data estimated by StabLM-Zephyr-3B. This indicates that different prior distributions yielded the same result, which is consistent with Figure~\ref{fig:ppl_distribution_llama_8b_lm}. The synthetic data lacks a long tail and is concentrated within a narrow portion of the distribution.} 
  \label{fig:ppl_StabLM-Zephyr-3B}
\end{figure*}

\section{Discussion}\label{sec:discussion}

\subsection{Non-Iterative vs Iterative Model Collapse.}\label{sec:Non-Iterative-Model-Collapse}
We define \textit{non-iterative model collapse} as the performance degradation caused by directly mixing general synthetic data with human-produced data, without iterative training. Theoretically, without additional regularization constraints to guide data generation, the variance of the model-generated data gradually decreases during this process. The diversity of the generated data diminishes over time, ultimately leading to the collapse of the model itself.

The difference between the two lies in their scope. Non-iterative model collapse is not confined to training on self-generated data, allowing it to uncover broader properties of synthetic data. For instance, in our experiments, we train GPT-2 on the Cosmopedia dataset in a single generation, which was generated by \texttt{Mixtral-8x7B-Instruct-v0.1}. In contrast, iterative model collapse focuses on training the model over multiple generations using self-generated data.

Furthermore, the non-iterative model collapse emphasizes the gap between human data and general purely synthetic data, particularly regarding distributional properties and n-gram features. In contrast, the iterative model collapse illustrates the iterative evolution of the model, resembling a self-play process.
This process illustrates the gradual evolution of self-generated data. It does not involve an analysis of the differences in nature between self-generated and human-produced data.
They both ultimately lead to model collapse, driven by the same underlying cause—synthetic data, though they investigate different aspects of synthetic data.
We often face that training a model on a mixture of human and synthetic data, where the synthetic data is not generated by the model itself, and its exact origin may be unknown. 

\subsection{Why Does the Observed Probability Distribution Exhibit Filtering Potential?}\label{sec:u_shape_distribution}
\textbf{From the perspective of information theory,} we can analyze the filtering potential of the U-shape distribution as follows: We utilize the U-shape distribution in Figure~\ref{fig:qwen2_probs} to re-sample tokens in the high-probability region, to adjust the U-shaped distribution toward a uniform distribution. By doing so, we can maximize the information entropy. According to information theory, maximizing information entropy is achieved when the distribution is uniform.

\textbf{Lemma 1}: Let \(X\) be a discrete random variable with \(n\) possible outcomes. If the probability of each outcome is uniform, i.e., \(P(x_i) = \frac{1}{n}\) for all \(i \in \{1, 2, \dots, n\}\), the Shannon entropy is maximized, given by:
\begin{align}
H(X) &= -\sum_{i=1}^{n} \frac{1}{n} \log \frac{1}{n} = \log n.
\end{align}
This represents the maximum uncertainty achievable, implying that the dataset carries the maximum possible information content. 
Thus, the uniform distribution, which assigns equal probability to all outcomes, possesses the maximum information entropy. To leverage this property, we utilize the U-shape distribution to re-sample tokens in the high-probability region, adjusting the U-shaped distribution toward a uniform distribution. By doing so, we can maximize the information entropy.

\textbf{From the perspective of language model learning,} our method emphasizes the importance of poorly learned data. Specifically, we resample easy tokens and encourage the model to focus on learning more challenging ones.
Our method can enhance the learning of underrepresented data by resampling high-probability tokens.

\subsection{Gradual Decline in Editing}\label{sec:Gradual_Decline_in_Editing}
\begin{table}[th!]
\centering
\caption{Percentage of tokens requiring edits in the Natural-Instructions dataset. The total number of tokens is 4,671,834.}
\label{tab:token_edits}
\begin{tabular}{@{}lccc@{}}
\toprule
\textbf{}               & \textbf{Gen 1 (source)} & \textbf{Gen 2} & \textbf{Gen 3} \\ \midrule
\textbf{Tokens ($p>0.99$)} & 584,103                      & 549,519               & 517,433               \\
\textbf{Percentage}        & 12.5\%                       & 11.76\%               & 11.08\%               \\ \bottomrule
\end{tabular}
\end{table}

We present the percentage statistics of edited tokens in Table~\ref{tab:token_edits} and performance in an iterative process in Table~\ref{tab:iterative_results}, demonstrating that the edited tokens indeed exhibit a progressive decrease. Specifically, We observe that the percentage of edited tokens (above the threshold $p>0.99$) decreases as the generation number increases. 
Theoretically,  this is a process of distribution shifting. When tokens ($p>0.99$) are resampled, randomness is introduced. The sampling process can select tokens with lower probabilities. Then, tokens ($p>0.99$) is replaced, leading to a reduction of edited tokens in subsequent generations. The Table~\ref{tab:token_edits} provides empirical evidence for this pattern of decay.

\begin{table}[h]
\centering
\caption{Performance in an iterative process on Instruction tuning data.}\label{tab:iterative_results}
\begin{tabular}{c|ccccc|c}
\toprule
       & PIQA  & BoolQ & HS    & SIQA  & WG    & Avg       \\
\midrule
Gen 0 & 79.87 & 81.28 & 59.72 & 49.69 & 74.51 & \textbf{69.01} \\
Gen 1 & 80.25 & 81.16 & 59.74 & 50.56 & 74.59 & \textbf{69.26} \\
Gen 2 & 80.14 & 82.69 & 59.82 & 50.51 & 73.80 & \textbf{69.39} \\
\bottomrule
\end{tabular}
\end{table}

\subsection{What is Coverage Narrowing?}
`coverage narrowing' refers to a phenomenon in which the distribution of synthetic data covers a significantly narrower range of values compared to human data, even when the data sizes are identical. For instance, as shown in Figure~\ref{fig:ppl_distribution_llama_8b_lm}, the PPL range of synthetic data is limited to $[0, 14]$, whereas the PPL range of human data extends from $[0, 100+]$. Despite this disparity, the overall coverage, represented by the area under the distribution curves, remains the same. This significant distribution gap is what we define as `coverage narrowing.'

\subsection{How Does the DSIR Work?}\label{sec:dsir}
DSIR~\citep{xie2023data} works by estimating importance weights for each data sample to measure its relevance to the target distribution. 
This involves three main steps: first, we leverage n-gram models to estimate two distributions of human and synthetic data, $q_{feat}$ and $p_{feat}$, which represent the target and raw distributions, respectively. We use them to compute the likelihood ratio for each sample. Next, we calculate the importance weight for each sample $z_i$ as $w_i = \frac{\hat{p}_{\text{feat}}(z_i)}{\hat{q}_{\text{feat}}(z_i)}$. The weight $w_i$ quantifies how well the sample aligns with the target distribution.  Finally, we perform importance-weighted sampling without replacement to select examples, ensuring that the selected data is more representative of the target distribution.

\begin{figure}
    \centering
    \includegraphics[width=\linewidth]{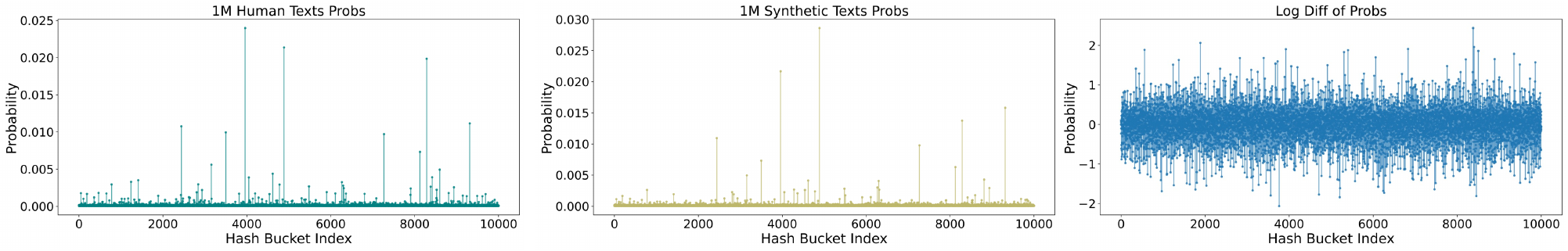}
    \caption{Uni/Bi-gram feature distribution across 10,000 hash buckets.}
    \label{fig:log_feature_diffence}
\end{figure}

We use DSIR in our data analysis as it allows for principled and computationally efficient selection of synthetic data points that align with the target distribution. Moreover, the importance weight also reflects the alignment between the n-gram features of synthetic data and human data. Using DSIR, we can analyze the differences between synthetic and human data across n-gram feature distributions and data matching. As shown in Figure~\ref{fig:log_feature_diffence}, it is challenging to select synthetic data that matches human data characteristics under the significant distribution difference. To obtain high-quality synthetic data, it is essential to focus on improving the data synthesis methods.

\subsection{Non-autoregressive Token Replacement May Compromise Text Coherence.}
When designing data synthesis algorithms, we must balance synthesis efficiency and effectiveness, considering both autoregressive and non-autoregressive approaches. Autoregressive methods leverage the inherent capabilities of language models to generate coherent text sequentially. In contrast, non-autoregressive methods resample individual tokens based on their probability distributions. Since data synthesis is a prerequisite for model training, we aim to ensure that the cost of data synthesis does not exceed the cost of training itself.

Specifically, our ToEdit modifies data using the probability distribution in a single forward pass. For instance, if the generated sequence length is 1024, the computational cost of autoregressive methods would be 1024 times higher than ours. This efficiency advantage is why our method can run effectively on GPUs like the 3090 or 4090 series.

However, this efficiency may come at the cost of coherence, as resampled tokens may not fit seamlessly into a given sentence. To address this issue, we introduce a hyperparameter, resampling probability $p$, to control the resampling threshold. We perform sampling in high-probability regions, focusing on tokens that are relatively easier to predict. We manually verify and tune on a small validation set before applying it across all experiments. In our experiments, we set $p=0.99$. 

Additionally, we supplement more experiments and discussion about hyper-parameter $p$. As Table~\ref{tab:ablation_resampled_p} shows, different values of $p$ influence BioMed performance, leading to fluctuations in data quality. Table~\ref{tab:token_distribution} presents the distribution percentages of the token probabilities across different value ranges. We need to refine the data while primarily preserving the source distribution. A larger $p$ indicates fewer tokens will be resampled, while a smaller $p$ results in more tokens being resampled. Balancing performance and the preservation of data distribution, we set $p=0.99$ as the threshold for our experiments.
\begin{table}[t]
  \centering
  \caption{PPL results of GPT-2 124M pretraining on pure Human or Synthetic data.}
  \resizebox{0.8\textwidth}{!}{
    \begin{tabular}{l|lllll|lllll}
\toprule
Data Type & \multicolumn{5}{c|}{Human Data (Dolma)} & \multicolumn{5}{c}{Synthetic Data (Cosmopedia)} \\
\midrule
Tokens Size & 8.4B & 16.8B & 25.2B & 33.6B & 42B & 8.4B & 16.8B & 25.2B & 33.6B & 42B  \\
\midrule
Epochs & 1 & 2 & 3 & 4 & 5 & 1 & 2 & 3 & 4 & 5 \\
\midrule
Wikitext-103 & 43.62 &  38.57 & 36.11 & 34.89 & 34.55 & 169.38 & 147.73 & 135.23 & 131.78 & 128.05 \\
RedPajama & 40.18 & 35.84 &  33.97 & 32.74 & 32.34 & 116.37 & 103.25 & 99.27 & 96.81 & 96.03 \\
Falcon-RefinedWeb & 54.85 &  49.10 & 46.93 & 45.43 & 44.90 & 146.97 & 132.60 & 127.68 & 124.32 & 122.69 \\
c4-en & 45.87 & 41.00 & 39.10 & 37.95 & 37.56 & 128.25 & 114.41 & 109.73 & 107.53 & 106.55 \\
mc4-en & 61.00 & 54.44 & 52.11 & 50.38 & 49.74 & 171.44 & 153.70 & 150.28 & 145.44 & 144.99 \\
\bottomrule
\end{tabular}}
\label{tab:gpt2_124m_pure_data}
\end{table} 

\subsection{Must We Assume the Data is 100\% Human-authored?}
We do not need to assume that the data is 100\% human authored; In experimental settings, some datasets used in our experiments include partially synthetic data:

\begin{itemize}
\item Datasets used in continual pretraining (e.g., Biomed, Finance) include partially  synthetic data, which has been reformatted into a reading comprehension structure~\citep{cheng2024adapting}.
\item OSS-Instruct-75K and Evol-Instruct-110K also contain samples synthesized by ChatGPT.
\end{itemize}
In the theoretical framework, synthetic data is generated iteratively through an $n$-generation process.  (1) If the starting point is a real distribution, our method preserves most of the initial distribution to generate higher-quality data. (2) If the starting point is a mixture of synthetic and real data, the modifications are minimal, ensuring the original distribution remains largely unaffected. Therefore, applying our method in any generation $i$, we can further avoid issues, such as reduced variance and diminished diversity,  which are key factors contributing to model collapse. 

In other words, whether the current data is fully real or a mix of real and synthetic, using it as anchor data to synthesize data, our method builds upon the current data distribution to achieve improvements, rather than causing model collapse.

In summary, we aim to improve the data synthesis method, specifically focusing on how to obtain higher-quality data from the existing datasets. We do not need to assume that the data at hand is 100\% human-generated. Our algorithm is designed to minimize excessive distribution truncation of the original data. 

\subsection{Can ToEdit Help with Already Strongly Collapsed Data or is a Minimum Quality of the Data Necessary?}
The ToEdit algorithm was initially designed to preserve the long-tail distribution during the data generation process, thereby avoiding model collapse. For already collapsed data, the variance is typically very small, and enhancing diversity is crucial. We can also adjust the threshold 
 to introduce more randomness in data. Through this operation, we can inject randomness into the collapsed data. However, this is a theoretical scenario, and as we know, data situations are highly complex. In practice, there will be many more challenges to address.

\subsection{Could We Generate Tons of Data Using LLMs and Select Only the Long-Tail Ones?}
For now, generating long-tail samples is currently difficult for language models. The reason lies in the sampling strategy of LLMs. Current LLMs adopt top-p, top-k or other sampling strategy for better performance. However, these sampling strategy will lead to cut-off distribution. When the data synthesizing scale up, this drawback will finally scaling law cut-off on synthetic data. However, human corpus data follows a Zipf distribution. The truncated output distribution causes the LLMs to nearly fail to sample long-tail samples. In other words, it is currently difficult to induce long-tail samples from LLMs that are as diverse as human data.

On the other hand, if we force the language model to generate long-tail samples, these may contain both noisy and high-information samples, which are like two sides of a coin, both distributed in the long tail of the data. This necessitates further filtering of the high-information samples. Unfortunately, such samples are challenging to automatically identify in practice and may require extensive human annotation.

\section{Potential Applications and Future Work}
Based on the above discussion, 
our approach can be applied to optimize the current data, even if it is a mixture of real and synthetic data. From the findings and proposed method in our paper, we can influence future research in the following aspects:

\paragraph{Potential applications of our work:} (1) Data optimizations. We can quickly modify and optimize the current data, using a trained language model with a single forward pass. (2) Regularization in the data synthesizing process. When synthetic data becomes excessive, we can introduce real data as an anchor to balance the issues of excessive homogeneity and tail distribution cut-off in synthetic data, thereby preventing mode collapse.



\paragraph{Lessons from our work:} The key to improving the quality of synthetic data lies in balancing long-tail distribution preservation and optimizing synthetic data approaches. In other words, we should focus on two questions: how to generate more informative synthetic data and how to integrate it with real data effectively. Building on this foundation, future improvements can focus on two aspects: first, obtaining more information gain by designing more efficient generation mechanisms to inject valuable information into the synthetic data; and second, optimizing methods to reduce noise during the synthesis process. This approach ensures that synthetic data retains its authenticity while enhancing its utility in practical tasks.

\paragraph{Extended related applications} A broader range of recent works across domains also explore synthetic data generation and usage for diverse applications~\cite{bai2025rat,bai2024efficient,Bai_Zhang_Tao_Wu_Wang_Xu_2023,jia2024bench2drive,yang2023llm4drive,li2020face,niu2024lightzero,zhang2023gobigger,li2024think2drive,pu2024unizero,li2023normalization,zhu2024critical,zhu-etal-2024-pad,zhu2022storytrans}.
\begin{figure}[h]
  \centering
  \includegraphics[width=\linewidth]{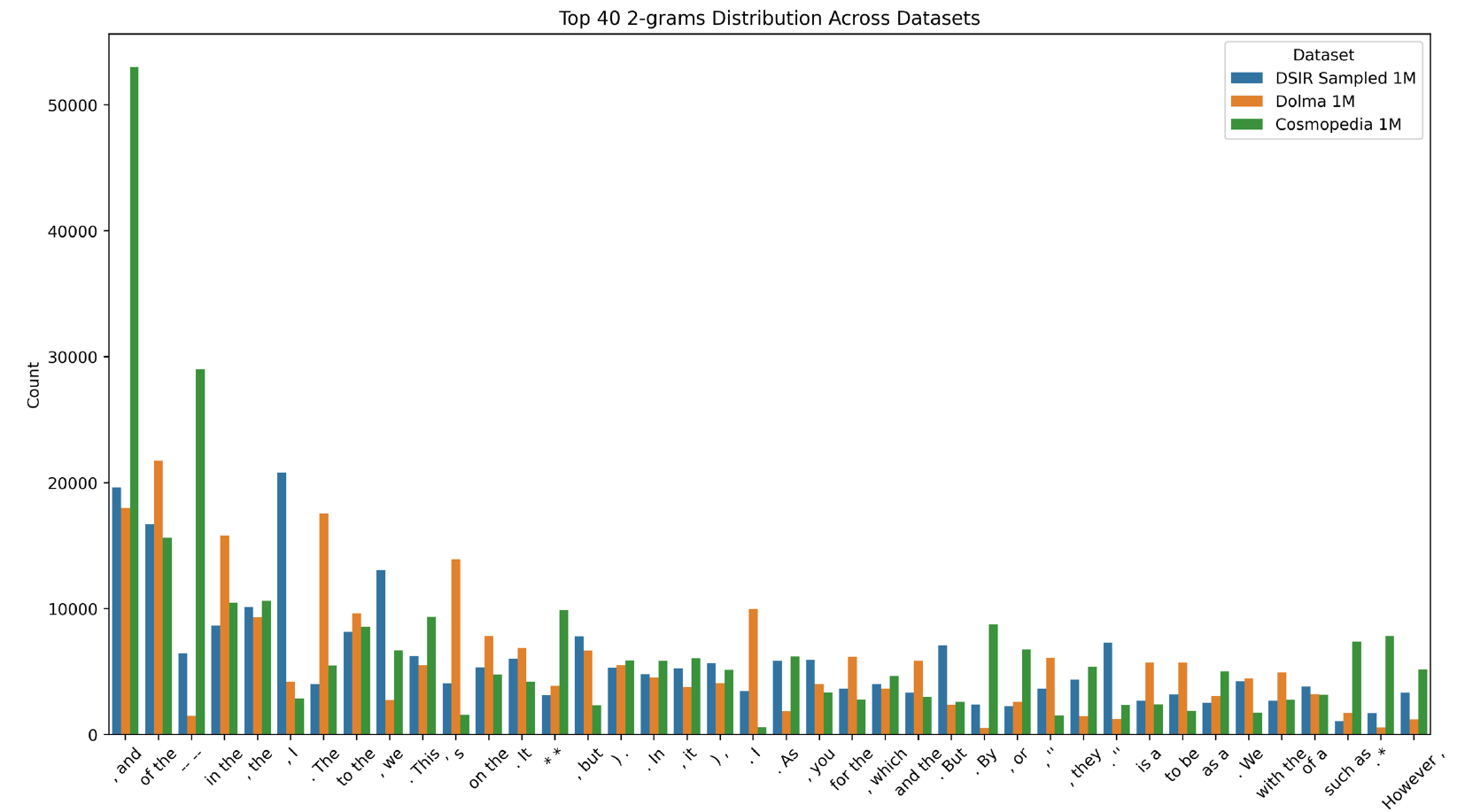}
  \caption{The top 40 bi-grams from separately sampled 1M subsets of Dolma, Cosmopedia, and DSIR-selected datasets.} 
  \label{fig:top_40_n_gram}
\end{figure}

\begin{figure*}[h]
  \centering
  \includegraphics[width=\linewidth]{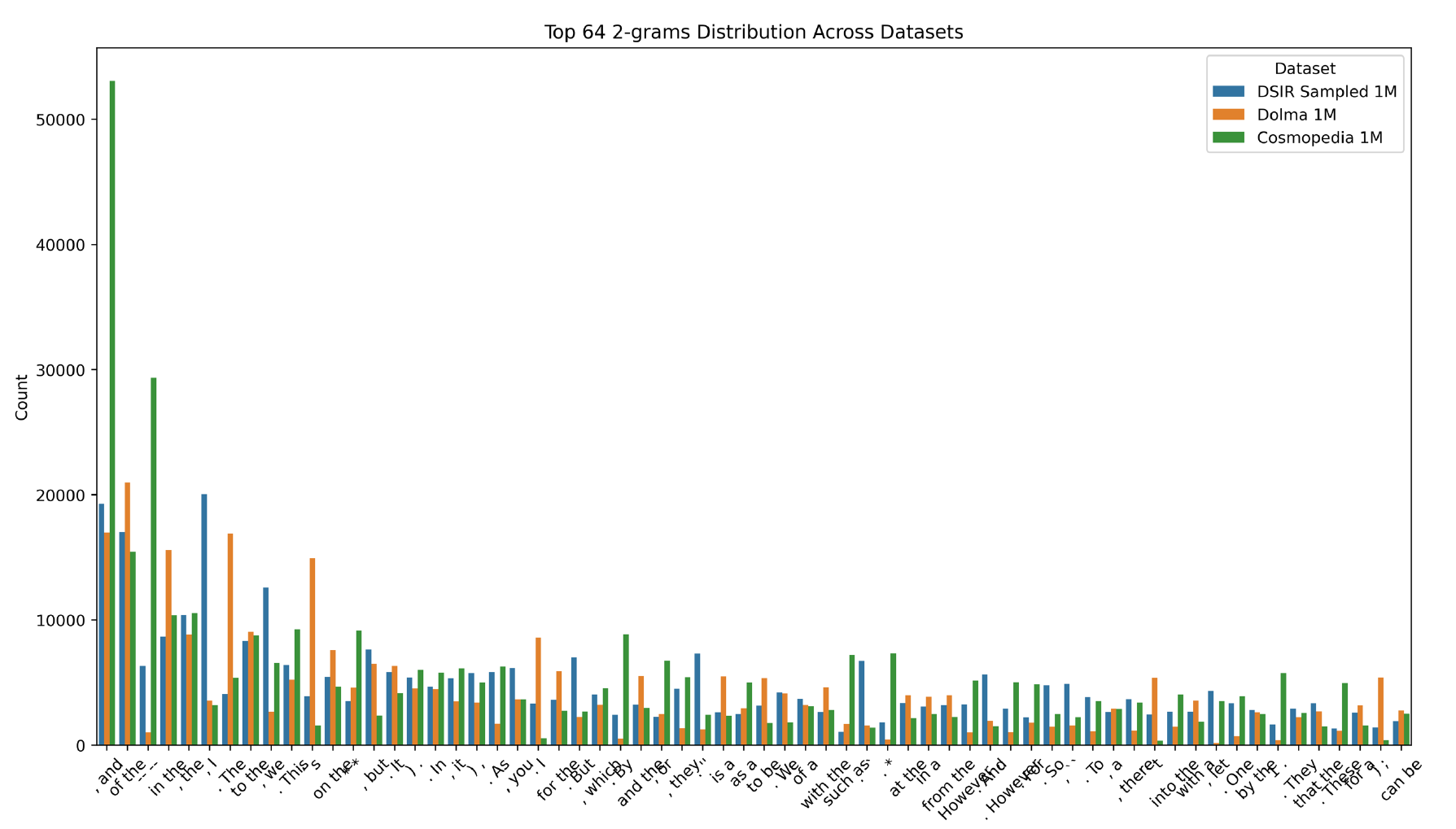}
  \caption{The top 64 bi-grams from separately sampled 1M subsets of Dolma, Cosmopedia, and DSIR-selected datasets.} 
  \label{fig:top_64_n_gram}
\end{figure*}

\begin{figure*}[h]
  \centering
  \includegraphics[width=\linewidth]{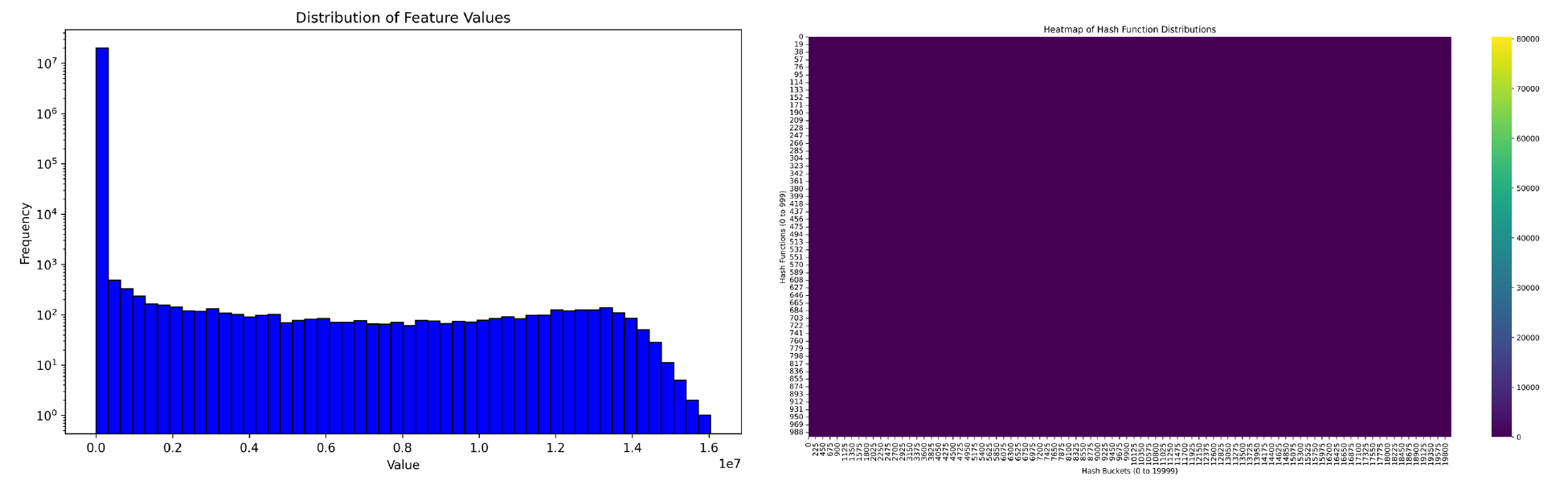}
  \caption{Density sampling response values. This result further confirms the issue of feature collapse in synthetic data.} 
  \label{fig:density_sampling}
\end{figure*}


\begin{table}[ht]
\centering
\caption{Dolma dataset statistics (v1.6), quoted from source~\citep{dolma}.}
\label{tab:data_sources}
\resizebox{\textwidth}{!}{
\begin{tabular}{@{}lccccc@{}}
\toprule
\textbf{Source}              & \textbf{Doc Type}     & \textbf{UTF-8 bytes (GB)} & \textbf{Documents (millions)} & \textbf{Unicode words (billions)} & \textbf{Llama tokens (billions)} \\ \midrule
Common Crawl                 & web pages             & 9,022                     & 3,370                          & 1,775                              & 2,281                            \\
The Stack                    & code                  & 1,043                     & 210                            & 260                                & 411                              \\
C4                           & web pages             & 790                       & 364                            & 153                                & 198                              \\
Reddit                       & social media          & 339                       & 377                            & 72                                 & 89                               \\
PeS2o                        & STEM papers           & 268                       & 38.8                           & 50                                 & 70                               \\
Project Gutenberg            & books                 & 20.4                      & 0.056                          & 4.0                                & 6.0                              \\
Wikipedia, Wikibooks         & encyclopedic          & 16.2                      & 6.2                            & 3.7                                & 4.3                              \\ \midrule
\textbf{Total}               &                       & \textbf{11,519}           & \textbf{4,367}                 & \textbf{2,318}                     & \textbf{3,059}                   \\ \bottomrule
\end{tabular}}
\end{table}




\end{document}